%% file: main.tex
\documentclass{article} 
\usepackage{style,times}

\input{math_commands.tex}

\usepackage{hyperref}
\usepackage{url}

\usepackage{tabularx}
\usepackage{booktabs}  
\usepackage{array}    

\usepackage{wrapfig}

\usepackage{graphicx}
\usepackage{xcolor}
\usepackage{enumitem, eucal}
\usepackage{hyperref}
\hypersetup{
    colorlinks=true,
    linkcolor=blue,
    citecolor=blue,
    urlcolor=blue,
    filecolor=blue,
}

\definecolor{mygreen}{rgb}{0, 0.5, 0}
\newcommand{\ours}{{\sc Ti\-nk\-er}}
\newcommand{\oursbf}{{\bfseries\scshape Ti\-nk\-er}}

\title{\oursbf: Diffusion's Gift to 3D
---Multi-View Consistent Editing From Sparse Inputs without Per-Scene Optimization}

\author{Canyu Zhao$^{1}$\thanks{Equal Contribution}
~~
Xiaoman Li$^{1*}$
~~
Tianjian Feng$^{1}$
~~
Zhiyue Zhao$^{1}$
~~
\textbf{Hao Chen}$^{1}$
~~
\textbf{Chunhua Shen}$^{1,2}$
\\[0.253cm]
$^1$ Zhejiang University, China~~~~
$^2$ Zhejiang University of Technology, China
}

\renewcommand{\vspace}[1]{} 

\iclrfinalcopy 
\begin{document}

\maketitle

\begin{abstract}

We introduce \ours, a versatile framework for high-fidelity 3D editing that operates in both one-shot and few-shot regimes without any per-scene finetuning. Unlike prior techniques that demand extensive per-scene optimization to ensure multi-view consistency or to produce dozens of consistent edited input views, \ours\  delivers robust, multi-view consistent edits from as few as one or two images. This capability stems from repurposing pretrained diffusion models, which unlocks their latent 3D awareness.

To drive research in this space, we curate the first large-scale multi-view editing dataset and data pipeline, spanning diverse scenes and styles. Building on this dataset, we develop our framework capable of generating multi-view consistent edited views without per-scene training, which consists of two novel components:
(1) 
Referring multi-view editor: Enables precise, reference-driven edits that remain coherent across all viewpoints.
(2) 
Any-view-to-video synthesizer: Leverages spatial-temporal priors from video diffusion to perform high-quality scene completion and novel-view generation even from sparse inputs.
Through extensive experiments, \ours\  significantly reduces the barrier to generalizable 3D content creation, achieving state-of-the-art performance on editing, novel-view synthesis, and rendering enhancement tasks. We believe that \ours\  represents a key step towards truly scalable, zero-shot 3D editing.
Project webpage:
 \href{https://aim-uofa.github.io/Tinker}{\oursbf}.

\end{abstract}

\input{tinker/Chapters/1_Intro}
\input{tinker/Chapters/2_Related_Works}
\input{tinker/Chapters/3_Method}
\input{tinker/Chapters/4_Experiments}
\input{tinker/Chapters/5_Conclusion}

\clearpage

\bibliography{ref}

\bibliographystyle{ref}

\clearpage
\appendix
\renewcommand\thesection{\Alph{section}}
\renewcommand\thefigure{S\arabic{figure}}
\renewcommand\thetable{S\arabic{table}}
\renewcommand\theequation{S\arabic{equation}}
\setcounter{figure}{0}
\setcounter{table}{0}
\setcounter{equation}{0}
\input{tinker/Appendix/Additional_Implementations}
\input{tinker/Appendix/Additional_Results}
\input{tinker/Appendix/Limitations}

\end{document}

%% file: math_commands.tex
\usepackage{amsmath,amsfonts,bm}

\def\eqref#1{equation~\ref{#1}}

\def\1{\bm{1}}

\DeclareMathAlphabet{\mathsfit}{\encodingdefault}{\sfdefault}{m}{sl}
\SetMathAlphabet{\mathsfit}{bold}{\encodingdefault}{\sfdefault}{bx}{n}



%% file: tinker/Chapters/1_Intro.tex
\begin{figure*}[htbp]
    \centering
    \includegraphics[width=\textwidth]{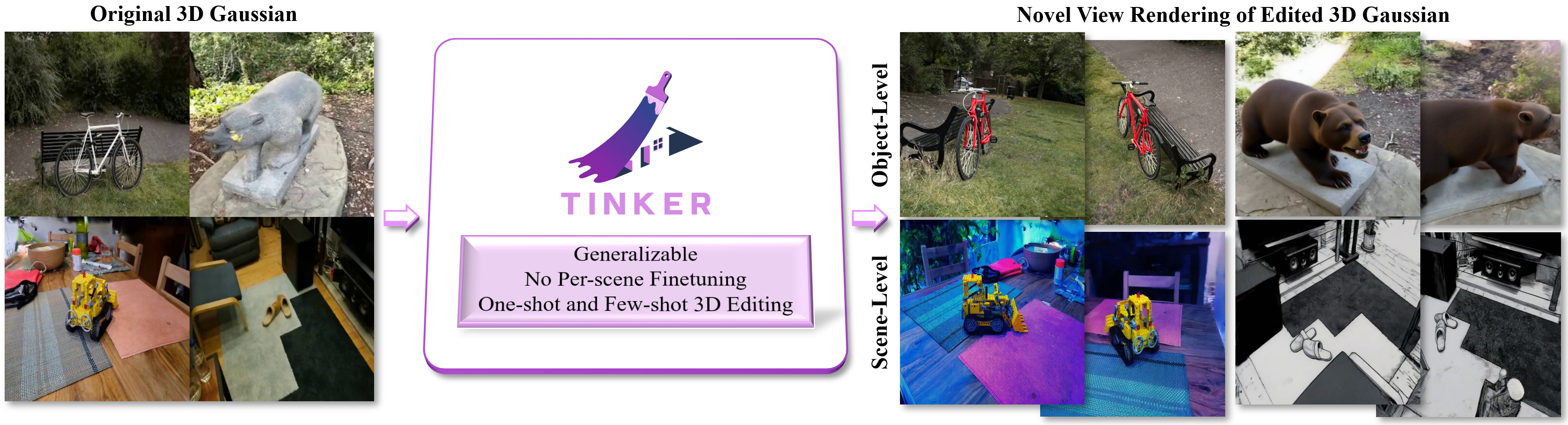}
    \vspace{-3mm}
    \caption{
        Compared with prior 3D editing approaches, \ours\ removes the necessity of labor-intensive per-scene fine-tuning--—whether for generating multi-view-consistent edited inputs for 3DGS optimization or for preserving consistency through scene-specific training.Moreover, \ours\ is capable of performing both object-level and scene-level 3D editing, and achieves high-quality results in few-shot as well as one-shot settings. \textbf{
        Please 
        refer to Figures \ref{fig:oneshot1}, \ref{fig:oneshot2}, \ref{fig:fewshot1}, \ref{fig:fewshot2} for more compelling visualizations.}
    }
    \label{fig:teaser}
    \vspace{-2mm}
\end{figure*}

\section{Introduction}

Benefiting from the rapid advancements in 2D diffusion models~\citep{rombach2022high,esser2024scaling,flux2024}, a prevailing paradigm for 3D editing has emerged: generating multi-view consistent images via 2D diffusion-based editing, followed by fine-tuning 3D Gaussian Splatting (3DGS)~\citep{kerbl20233d} or Neural Radiance Field (NeRF)~\citep{mildenhall2021nerf} to edit 3D scenes. This pipeline has become the de facto standard in recent 3D editing approaches.

\begin{figure*}[tb]
    \vspace{-2mm}
    \centering
    \includegraphics[width=\textwidth]{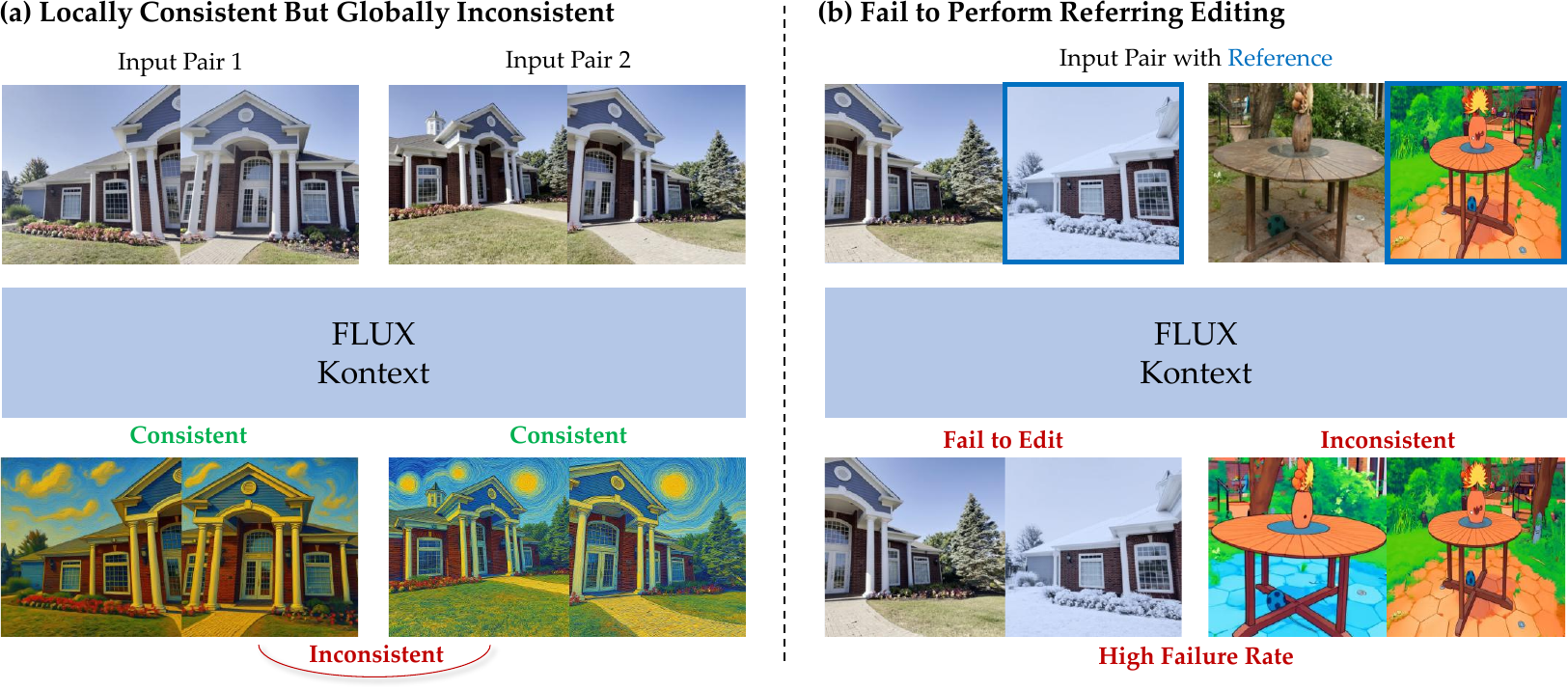}
    \vspace{-5mm}
    \caption{
    (a) FLUX Kontext achieves multi-view consistent image editing by horizontally concatenating two images and editing them jointly. Although it ensures consistency between the concatenated image pair, significant inconsistencies remain across different image pairs.
    (b) We also evaluated whether FLUX Kontext can edit one half of the concatenated image by referencing the other half. The results demonstrate that the model lacks this capability.
    }
    \label{fig:mv_edit}
    \vspace{-2mm}
\end{figure*}
During the era when U-Net-based diffusion models dominated, many successful 3D editing approaches were inspired by advances in 2D image editing, such as Instruct-NeRF2NeRF~\citep{haque2023instruct} and InstructPix2Pix~\citep{brooks2023instructpix2pix}.
Recently, the emergence of Diffusion Transformer (DiT) architectures~\citep{peebles2023scalable} and Flow Matching~\citep{liu2022flow,lipman2022flow,albergo2022building,esser2024scaling} has significantly advanced the field of generative modeling. Latest developments have demonstrated substantial improvements in both image and video generation~\citep{wan2025wan,esser2024scaling,flux2024}, editing~\citep{wang2024taming,yu2025anyedit,ku2024anyv2v,jiang2025vace,labs2025flux1kontextflowmatching}, and even vision understanding~\citep{zhao2025diception,wang2025lavin,ke2024repurposing}, all driven by large-scale DiT flow-based models.
Theoretically, 3D editing should also naturally evolve to incorporate and benefit from these powerful new architectures and methodologies. However, we find that current 3D editing methods have yet to fully capitalize on these recent breakthroughs.
Many recent approaches, despite producing impressive results, remain constrained by conventional U-Net-based methodology~\citep{haque2023instruct,wu2024gaussctrl,zhuang2024tip,fujiwara2024style}, rather than embracing the more powerful and scalable techniques~\citep{wang2025personalize,zhang2025v2edit}.
There remains a noticeable disconnect between the progress made in 2D editing and the latest methods in 3D editing.
One key reason behind this limitation is the lack of multi-view consistent image editing datasets. For recent unified generation and editing models, fine-tuning with large-scale data has proven to be highly effective. However, the difficulty in collecting high-quality multi-view consistent datasets hinders progress in 3D-aware or view-consistent editing tasks.

Inspired by the remarkable capabilities of large language models in addressing unseen tasks~\citep{wei2021finetuned,achiam2023gpt,guo2025deepseek}, we pose a natural question: can recent large-scale image editing foundation model~\citep{labs2025flux1kontextflowmatching} also perform the multi-view consistent editing?
The answer is
confirmative.
We observe that simply concatenating two images as input enables these models to produce highly consistent and high-quality edits across views. However, while this pairwise concatenation ensures consistency between the two input views, we find that significant discrepancies often arise between different image pairs, thereby limiting global view consistency.
A straightforward idea is to concatenate an unedited image with an edited one, using the latter as a reference to guide the editing process. However, we observe that current foundation model does not exhibit this capability, usually producing inconsistent results, and typically reproduces the unedited image without modification. We illustrate both cases in Figure~\ref{fig:mv_edit}.

To address this issue, we design a novel pipeline that amplifies the model's capability for multi-view consistent editing. Specifically, we first introduce a data pipeline to generate referring editing dataset, where an unedited image is concatenated with another view that has already been edited. Fine-tuning with this dataset encourages the image editing model to learn how to propagate the editing intent across different viewpoints, which significantly improves the editing success rate and promotes better cross-view consistency.
Furthermore, to efficiently propagate edits from a sparse set of edited views to a dense set of novel views, we introduce a view completion model, which effectively bridges 2D and 3D editing by leveraging video editing. 
Unlike previous approaches that rely on repeated fine-tuning either to enforce multi-view consistency or to obtain multi-view-consistent input views for downstream 3DGS optimization, \textbf{the key distinction of \oursbf\ lies in its ability to directly produce high-quality, multi-view-consistent edited input views that can be seamlessly leveraged for 3DGS optimization without per-scene fine-tuning.}
Additionally, by fully exploiting the priors embedded in foundation models, our approach is also able to enhance the overall rendering quality of 3D scenes.
We believe \ours\ provides a new perspective on how foundation models can be adapted to 3D tasks and paves the way for future research in generalizable, user-friendly 3D content creation.

In summary, our main contributions are as follows:
\begin{itemize}[leftmargin=*]
\itemsep 0cm
    \item We design a novel pipeline that effectively elicits the multi-view consistent editing capabilities of large-scale generative models, and introduce, to the best of our knowledge, the first multi-view consistent image editing dataset.
    
    \item We introduce a sparse-view completion model specifically tailored for 3D editing tasks by rethinking editing problem as reconstruction problem. In addition to editing, our model is capable of performing video reconstruction.
    
    \item Our \ours\ eliminates the requirement for per-scene optimization that previous methods necessitate to ensure multi-view consistency or to generate multi-view-consistent edited input views, thereby significantly lowering the barrier 
    for practical use of 3D editing.  We hope that 
    \ours\ can serve as a general-purpose foundation for future advancements in 3D editing.
\end{itemize}

%% file: tinker/Chapters/2_Related_Works.tex
\section{Related Work}

\subsection{Diffusion Model}
Diffusion models~\citep{ho2020denoising,song2020denoising,rombach2022high,esser2024scaling,flux2024} are a powerful class of generative models that produce high-quality and diverse outputs by learning to reverse a gradual noising process. This process consists of a forward stage, where data is incrementally corrupted by Gaussian noise over multiple steps, and a reverse stage, where a neural network is trained to iteratively denoise and reconstruct the original data. 
Transformer-based architectures~\citep{peebles2023scalable,vaswani2017attention} and flow-matching~\citep{albergo2022building,lipman2022flow,liu2022flow} objectives have recently become the mainstream design choices in diffusion models~\citep{esser2024scaling,flux2024}, offering significant improvements in generation quality and scalability.
Owing to their strong generative priors acquired by large-scale training, diffusion models have significantly advanced a variety of vision tasks, such as image and video generation~\citep{blattmann2023stable,wan2025wan}, editing~\citep{brooks2023instructpix2pix,yu2025anyedit,labs2025flux1kontextflowmatching,tian2025mige}, image perception~\citep{ke2024repurposing,zhao2025diception,wang2025lavin}.

\subsection{2D Editing}
To achieve image editing, some pioneering studies explored alterations to the attention mechanism within the generative model~\citep{hertz2022prompt,chefer2023attend}.
A majority of approaches in image editing primarily revolved around inversion-based methods. These methods work by first inverting an input image back into its latent noise representation, and subsequently use a new prompt to generate the edited image~\citep{mokady2023null,cao2023masactrl,song2020denoising,wang2024taming,rout2024beyond}.
Beyond the inversion-based methods, some methods directly train models to follow explicit editing instructions~\citep{brooks2023instructpix2pix,pan2023kosmos}.
Similarly, the paradigm of video editing largely aligns with that of image editing~\citep{qin2024instructvid2vid,liew2023magicedit,liu2024video,geyer2023tokenflow,ku2024anyv2v,khachatryan2023text2video}.
Recently, high-quality unified models have emerged in both image~\citep{yu2025anyedit,labs2025flux1kontextflowmatching} and video editing~\citep{jiang2025vace}.
However, we observe that most mainstream approaches do not focus on multi-view consistent editing~\citep{liu2023zero,liu2023one}. While some recent methods~\citep{jiang2025vace} can perform depth-conditioned video editing, their primary focus lies in generation, which often results in videos with large motion and dynamics, causing multi-view inconsistencies.

\subsection{3D Editing}
3D Gaussian Splatting (3DGS)~\citep{kerbl20233d} and Neural Radiance Fields (NeRF)~\citep{mildenhall2021nerf} are two widely adopted 3D representations in recent years. 
Early approaches typically performed style transfer by learning a mapping between the source and target scenes~\citep{liu2023stylerf}.
With the rapid advancement of 2D diffusion models, numerous 3D editing methods have incorporated them as key modules~\citep{chen2024dge,fujiwara2024style,haque2023instruct,wu2024gaussctrl,zhuang2024tip}.
Some methods~\citep{chen2024shap,decatur20243d,dong2024interactive3d,sella2023vox} leverage Score Distillation Sampling (SDS)~\citep{poole2022dreamfusion} to perform editing by guiding the optimization of 3D representations with gradients derived from powerful pretrained diffusion models.
Nowadays, the prevailing paradigm is to leverage diffusion models to generate or edit a sufficient number of views, which are then used to finetune the underlying 3DGS.
However, while recent 2D diffusion models have seen significant breakthroughs in both generation quality and editing controllability, many of the latest 3D personalization approaches still rely on early U-Net-based architectures and approaches~\citep{kim2024dreamcatalyst,zhang20243ditscene}, thereby failing to take advantage of the latest advancements in the image and video diffusion models.
While a small number of recent approaches~\citep{wang2025personalize,zhang2025v2edit} have adopted state-of-the-art image and video diffusion models and demonstrated high-quality 3D editing results, they often depend on per-scene finetuning, which poses challenges in terms of efficiency and scalability.
In contrast, our method not only fully leverages recent developments in 2D diffusion models~\citep{labs2025flux1kontextflowmatching,wan2025wan}, but also eliminates the need for per-scene training, achieving more compelling 3D editing in a simple yet elegant manner.

%% file: tinker/Chapters/3_Method.tex
\section{Method}

We begin in Section~\ref{sec:3d_editing} by elucidating \ours's process for both few-shot and one-shot editing. We then provide a detailed account in Section~\ref{sec:image_editing} of how we constructed our dataset and model for multi-view consistent image editing. In Section~\ref{sec:scene_completion}, we introduce our model for scene completion from sparse views. Section ~\ref{sec:additional_feature} further discusses \ours's additional applications and potential.

\begin{figure*}[tb]
    \centering
    \includegraphics[width=\textwidth]{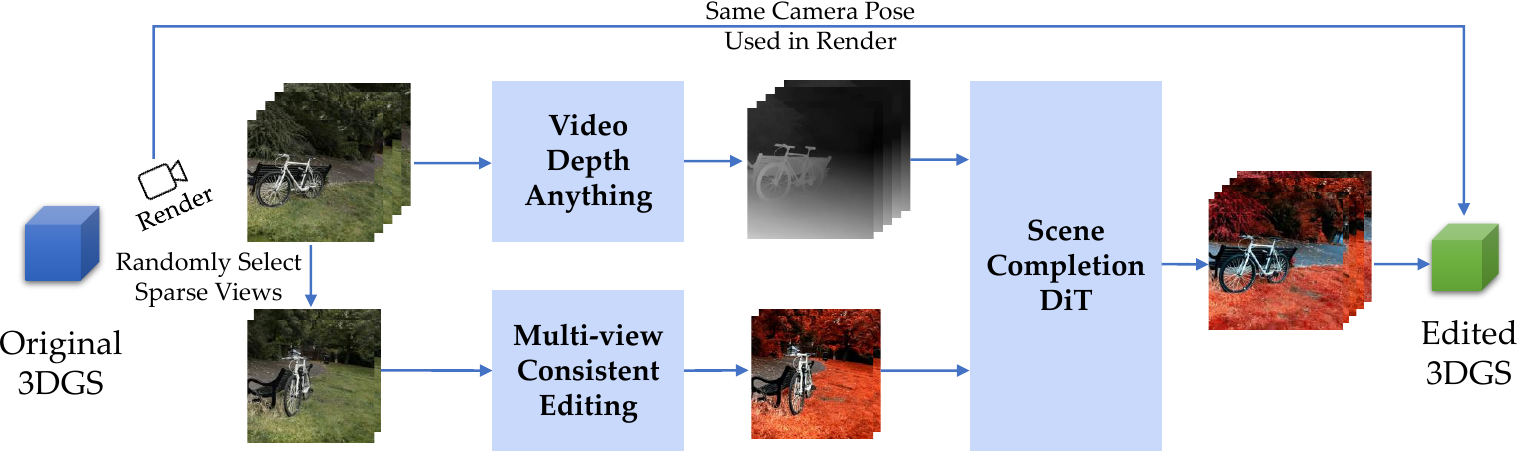}
    \vspace{-3mm}
    \caption{
    Overview of our editing process. We first apply our multi-view consistent editing model to obtain coherent sparse views. Leveraging depth constraints from the rendered results, we generate a large number of consistent edited images. The edited images are used to optimize the 3DGS to achieve high-quality 3D editing.
    }
    \label{fig:editing_process}
    \vspace{-2mm}
\end{figure*}

\subsection{3D Editing with Sparse Views as Input}
\label{sec:3d_editing}
\textbf{Editing with few-shot input.} 
As illustrated in Figure~\ref{fig:editing_process}, given an original 3DGS $\mathcal G$, our objective is to generate an edited 3DGS $\mathcal G'$. We begin by rendering a few videos from $\mathcal G$ and randomly selecting a few sparse views. These selected views are edited using our multi-view consistent image editing model to produce the edited reference views.
We then estimate the depth maps of the rendered video using Video Depth Anything~\citep{chen2025video}. With these depth maps and the edited reference views, we employ our scene completion model to generate the images of the other views.
Since the video is rendered from $\mathcal G$, we have access to the exact camera pose for each view, which allows the completed frames to be directly used in optimizing $\mathcal G$ into the edited $\mathcal G'$. The entire editing does not necessitate any per-scene finetuning.

\textbf{Editing with one-shot input.}
\ours\ is also capable of handling a more challenging scenario: editing with one single reference view without additional training. The procedure remains identical to the few-shot scenario, with the key difference being that we select only a single sparse view from the rendered videos to serve as the initial reference for editing.
This initial reference view is then used by the scene completion model to generate an initial set of edited views. These newly generated views, in turn, serve as subsequent reference views to progressively propagate the edit, with the process continuing until the entire scene is sufficiently covered by the generated views. As a result, edited $\mathcal G'$ is achieved by fine-tuning $\mathcal G$ with these generated views.

\subsection{Multi-view Consistent Image Editing}
\label{sec:image_editing}
Our approach begins with the observation that the state-of-the-art large-scale image editing model~\citep{labs2025flux1kontextflowmatching} is capable of achieving multi-view consistent editing when provided with two concatenated images as input. This simple strategy ensures consistency between the two concatenated views. However, it fails to enforce consistency across different image pairs, leading to global inconsistencies.
A straightforward solution is to concatenate an edited image with an unedited one, prompting the model to apply similar edits to the latter by implicitly referencing the former.
However, we find that this approach yields a very low success rate, with the model frequently reverting to simply reproducing the original unedited image. The model is not exposed to such reference-based editing configurations during its pretraining, and therefore lacks the knowledge needed to generalize in this setting.

\begin{figure*}[tb]
    \centering
    \includegraphics[width=\textwidth]{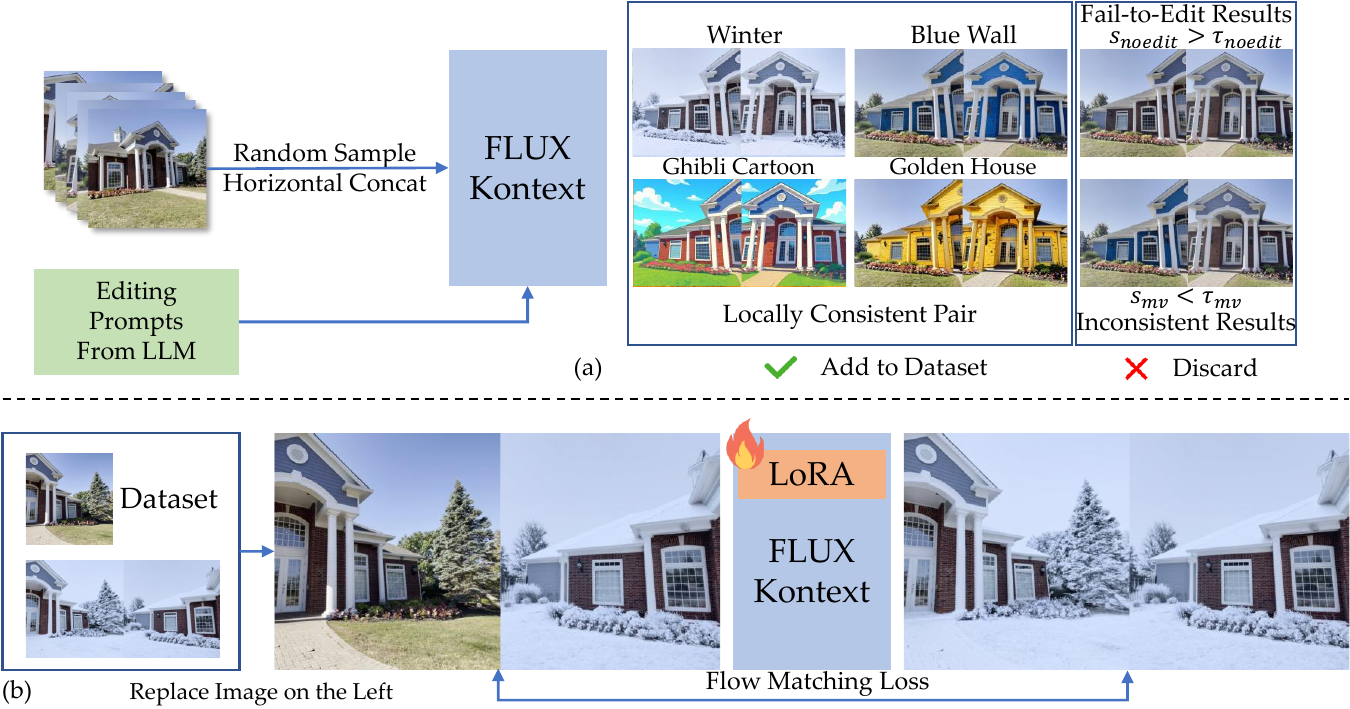}
    \vspace{-3mm}
    \caption{
        (a) We leverage the base FLUX Kontext model to generate a large number of consistent image pairs and discard the inconsistent and fail-to-edit results.
        (b) The data generated in (a) is used to fine-tune the model. Specifically, we horizontally concatenate the input image and an edited image, and train the model using LoRA to learn how to perform referring-based editing.
    }
    \label{fig:mv_edit_pipeline}
    \vspace{-2mm}
\end{figure*}

Inspired by the strong local consistency exhibited by the large image editing model, we leverage the model itself to construct a large-scale reference-based editing dataset.
Specifically, we begin by randomly selecting two different views of the same scene from publicly available 3D-aware datasets~\citep{yeshwanth2023scannet++,baruch2021arkitscenes,liu24uco3d,ling2024dl3dv,xia2024rgbd,zhou2018stereo}.
Given a 3D-aware dataset $\mathbb D=\{\mathbf I_v^i\}$, where $\mathbf I_v^i$ denotes the $i$-th scene from view $v$, we sample 2 views $\mathbf I_a, \mathbf I_b \in \mathbb D$ of the same scene from different viewpoints. 
We omit the scene index $i$ for simplicity to indicate that 2 views are sampled from the same scene.
Subsequently, we generate editing prompts $P$ with a large language model and perform editing using model $\mathcal E$: 
\begin{equation}
    \mathbf{I}'_a, \mathbf{I}'_b = \mathcal{E}(\operatorname{Concat}(\mathbf{I}_a, \mathbf{I}_b),P)
\end{equation}
To ensure that the editing is effectively applied, we compute the feature similarity between the edited and original images using DINOv2~\citep{oquab2023dinov2}:
\begin{equation}
    \begin{aligned}
        s_1 = \operatorname{sim}(f_{dino}(\mathbf I_a),f_{dino}(\mathbf I_a'))&, s_2 = \operatorname{sim}(f_{dino}(\mathbf I_b),f_{dino}(\mathbf I_b')), s_{noedit} = \operatorname{max}(s_1, s_2)
    \end{aligned}
\end{equation}
We discard pairs with overly high similarity scores that exceed a threshold $\tau_{noedit}$ by $s_{dino} >\tau_{noedit}$, indicating insufficient editing.
Furthermore, we evaluate the similarity between the two edited views by $s_{mv} = \operatorname{sim}(f_{dino}(\mathbf I'_a),f_{dino}(\mathbf I'_b))$, and filter out samples with low inter-view consistency below a threshold $\tau_{mv}$ by $s_{mv}<\tau_{mv}$.
Finally, we construct training inputs by pairing an original image with a reference edited image from a different view, and fine-tune the model using LoRA~\citep{hu2021lora} to learn reference-based editing. This process empowers the model to generalize edits across views and achieve globally consistent results. This process and the training objective can be formulated as:
\begin{equation}
    \begin{aligned}
    \mathbf I = \operatorname{Concat}(\mathbf I_a, \mathbf I'_b)&, \mathbf I' = \operatorname{Concat}(\mathbf I_a', \mathbf I'_b) \\
    \mathbf z_0 = g(\mathbf I) &,  \mathbf z' = g(\mathbf I') \\
        {\rm Loss } = 
        {\mathbb E}_{\mathbf z_0,t} \Vert \mathcal E_\theta&(\mathbf z_t,t,P) - u(\mathbf z'_t) \Vert^2_2,
    \end{aligned}
\end{equation}
where $g$ is the Variational Autoencoder that maps images to the latent space. We employ a flow matching loss~\citep{liu2022flow,lipman2022flow,albergo2022building,esser2024scaling} to minimize the discrepancy between the model's predicted velocities and the ground truth velocities $u$.

\subsection{Scene Completion Model}
\label{sec:scene_completion}
While it is feasible to perform view-by-view editing by leveraging our multi-view consistent image editing model using sparse reference views, this approach is extremely time-consuming. Motivated by the recent success of video diffusion models, we aim to exploit their strong spatial-temporal priors to achieve efficient scene completion with sparse edited views. A natural idea is to train a model that edits the original 3DGS rendered video into a target video, conditioned on multi-view consistent edited images. However, there are no such editing datasets currently available.
\textbf{Therefore, we rethink the problem by casting the training objective of the editing task into the reconstruction task.} Specifically, we aim to train a model to reconstruct the original scene from sparse views, with the goal of generalizing to reconstructing the edited scene from edited views, thus achieving editing through reconstruction.

\begin{figure*}[tb]
    \centering
    \includegraphics[width=\textwidth]{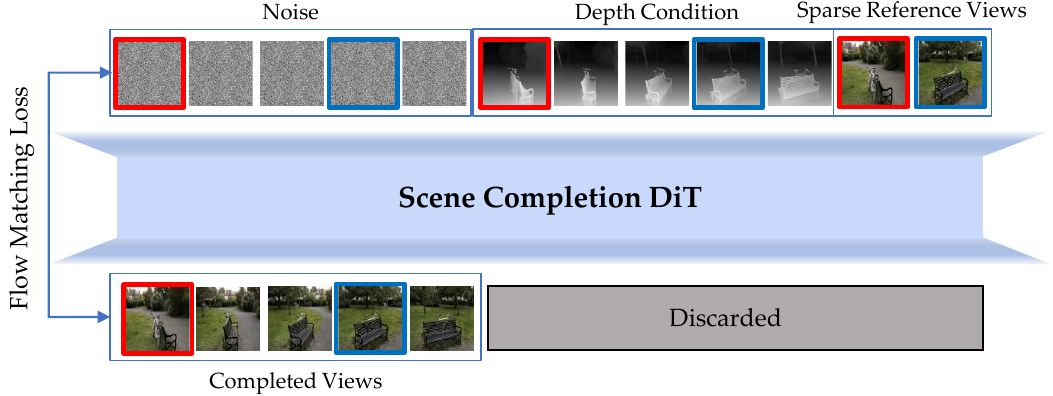}
    \vspace{-3mm}
    \caption{
    We build our Scene Completion Model on top of WAN2.1. In this figure, contours with the same color indicate the elements using the same positional embedding. During training, we compute the flow matching loss only between the model's outputs corresponding to the noisy latents and the target video.
     }
    \label{fig:view_completion}
    \vspace{-2mm}
\end{figure*}
In generative modeling, several works have attempted to condition multi-view generation on ray maps that encode camera parameters. However, we find that this approach lacks sufficient geometric constraints, often resulting in view inconsistencies.
Upon further analysis, we argue that depth is a more suitable conditioning signal in the context of 3D. Depth not only explicitly encodes structural constraints but also implicitly carries information about camera pose. Moreover, it greatly benefits from recent advances in depth estimation.

Recent unified model for video generation and editing, VACE~\citep{jiang2025vace}, has explored depth-guided video editing. However, as generation is their main objective, they treat depth more as a soft reference than an explicit constraint, resulting in results that deviate from the intended geometric structure. 
Although its performance is remarkable in depth-guided generation, this flexibility is undesirable in 3D editing scenarios where we require precise control over specific areas while maintaining geometric consistency in unedited regions.
3D editing necessitates a 3D-aware scene completion model that accurately follows depth constraints to maintain strict geometric consistency throughout the scene.

Based on the above analysis, we develop our scene completion model tailored for 3D. To this end, we leverage the pretrained WAN2.1~\citep{wan2025wan} to train an image-to-video model that accurately follows depth conditions. 
Given the scarcity of 3D video datasets, we take inspiration from Diception~\citep{zhao2025diception}, which achieves strong performance with limited data. 
Specifically, we treat the depth maps as RGB images and process them into tokens, following the original procedure of Wan2.1. The reference views are also processed into tokens in the same manner.
The noisy latent tokens $\mathbf Z^t=[\mathbf Z^t_0,\mathbf Z^t_1,\dots,\mathbf Z^t_N]$ at timestep $t$, depth tokens $\mathbf D=[\mathbf D_0,\mathbf D_1,\dots,\mathbf D_N]$, and reference view tokens $\mathbf V$ are concatenated along the sequence dimension to form the model input. The training process also follows the flow matching, which is formulated as:
\begin{equation}
    \begin{aligned}
&\mathbf X^t_{input} = \operatorname{Concat}(\mathbf Z^t, \mathbf D,\mathbf V) \\
    {\rm Loss } &= 
        {\mathbb E}_{\mathbf z_0,t} \Vert \Phi_\theta(\mathbf X^t_{input},t) - u(\mathbf Z^t) \Vert^2_2.
    \end{aligned}
\end{equation}
During training, we always provide the first frame as the default reference view, and randomly select 0 to 2 additional reference views to help the model learn both one-shot and few-shot scene completion.
The text embedding is fixed to a constant embedding, enhancing the model to focus solely on depth-guided generation.
To enable the model to associate these reference views with the target $j$-th frame, we assign them the same positional embedding as the target frames:
\begin{equation}
    \operatorname{PE}(\mathbf V) = \operatorname{PE}(\mathbf D_j) = \operatorname{PE}(\mathbf X_j).
\end{equation}
Through this design, the model effectively learns which regions of the scene correspond to the reference view, and is able to leverage both depth and reference views to achieve high-quality scene completion.

%% file: tinker/Chapters/4_Experiments.tex
\section{Experiments}

\subsection{Comparative Experiments}
\begin{figure*}[tb]
    \centering
    \includegraphics[width=\textwidth]{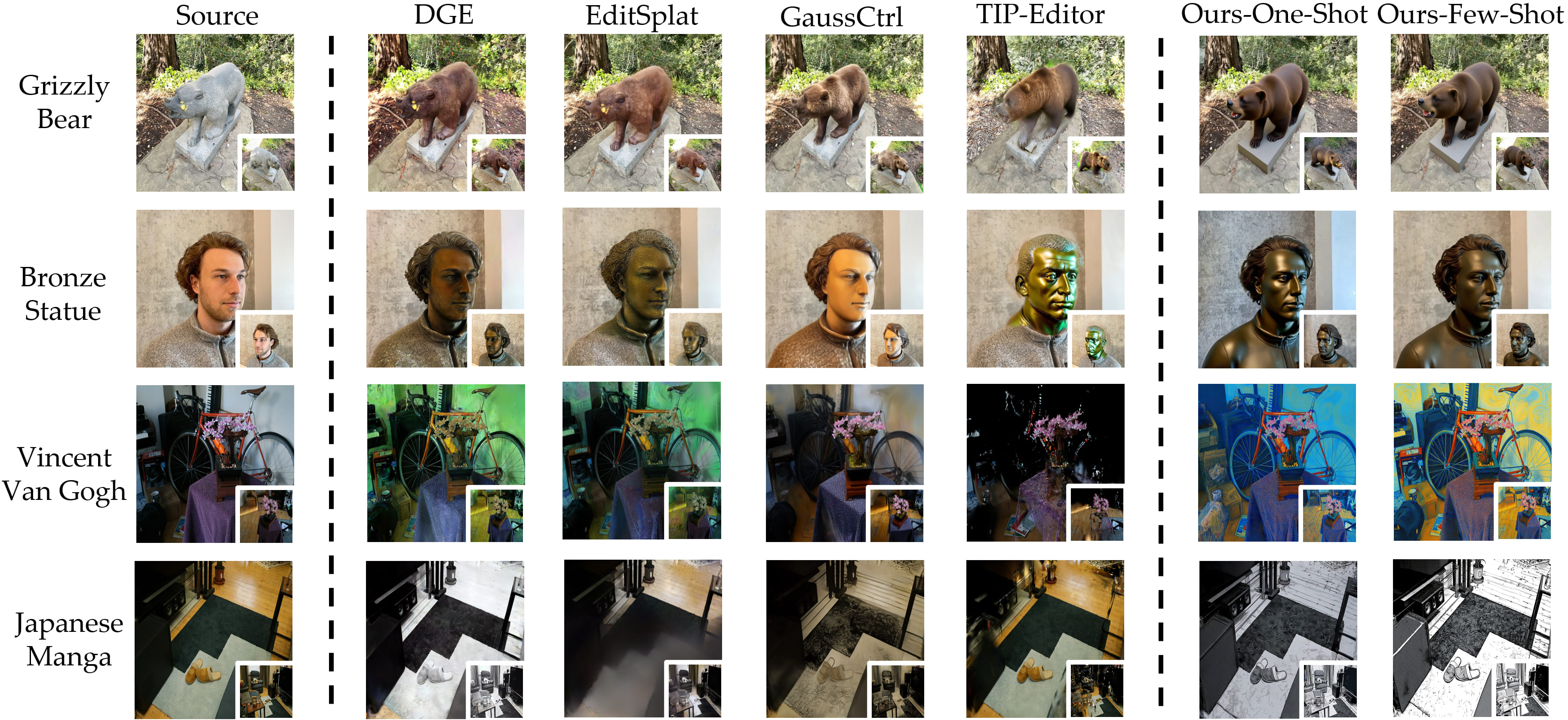}
    \vspace{-3mm}
    \caption{
    Qualitative comparisons of novel views in different methods.
    }
    \label{fig:qualitative_compare}
    \vspace{-2mm}
\end{figure*}
Due to space limitations, the implementation details are provided in Appendix~\ref{append:additional_implementations}. Building upon this implementation,
we conduct a comprehensive qualitative and quantitative comparison with the latest high-quality 3D editing methods~\citep{wu2024gaussctrl,chen2024dge,zhuang2024tip,lee2025editsplat} in terms of both output quality and computational cost. 
Our evaluation is conducted on the Mip-NeRF-360~\citep{barron2022mip} and IN2N~\citep{haque2023instruct} datasets. For each scene, we perform editing using the same textual prompt, and subsequently render the edited scenes from uniformly sampled camera poses to obtain results from different methods under identical novel viewpoints. For methods necessitating image input, we follow their official preprocessing pipelines to prepare the corresponding images. We use NeRFStudio~\citep{tancik2023nerfstudio} for 3DGS optimization and rendering. Furthermore, we quantitatively evaluate the results using four metrics: CLIP Text-Image
directional similarity~\citep{radford2021learning} used in GaussCtrl~\citep{wu2024gaussctrl} assess semantic alignment, DINO similarity~\citep{oquab2023dinov2} between the edited renderings to measure cross-view consistency, and aesthetic score~\citep{schuhmann2022laion5bopenlargescaledataset} to assess rendering quality.

\input{tinker/Tables/compare}
As demonstrated in Figure~\ref{fig:qualitative_compare} and Table~\ref{tab:compare}, in both one-shot and few-shot settings, our approach consistently outperforms existing methods for both object-level and scene-level editing. Furthermore, some methods, such as GaussCtrl~\citep{wu2024gaussctrl}, require per-scene fine-tuning, making them infeasible to run on consumer-grade 24 GPUs, whereas our method works entirely without further per-scene fine-tuning and can be executed efficiently on a single consumer-grade GPU. 
Moreover, we observe that while some approaches such as TIP-Editor~\citep{zhuang2024tip} are capable of producing high-quality object-level edits, they fall short in performing scene-level edits. In contrast, our method supports both object-level and scene-level 3D editing with higher quality, even for scenes with substantial overall style changes, such as oil paintings or black-and-white comics.

\subsection{Ablations and Analyses}
We performed a comprehensive set of ablation studies and analyses to better understand the effectiveness of our design choices. Specifically, we examined the effectiveness of fine-tuning our multi-view consistent editing model, assessed the effect of concatenating additional images for consistent editing, and evaluated the advantages of employing depth as a conditioning signal over the ray-map conditioning used in prior work. Moreover, we analyzed the strengths of our approach relative to existing depth-guided video generation methods. Owing to space constraints, the video-related ablation studies are shown in Appendix~\ref{append:additional_ablations}.

\begin{figure*}[tb]
    \centering
    \includegraphics[width=.95\textwidth]{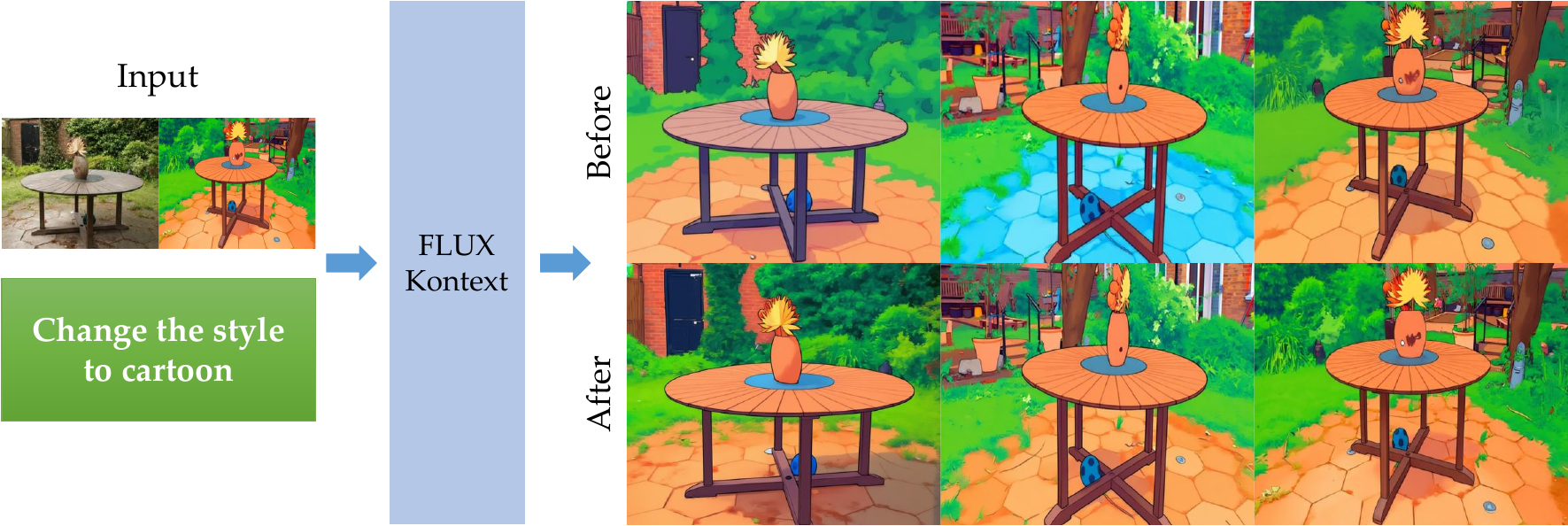}
    \vspace{-3mm}
    \caption{
    Qualitative comparisons before and after multi-view consistent image editing fine-tuning. After fine-tuning, our model can perform edits guided by the provided reference image, effectively ensuring the global consistency.
    }
    \label{fig:flux_finetune}
    \vspace{-2mm}
\end{figure*}
\textbf{Finetuning FLUX-Kontext for multi-view consistent editing.}
To validate the effectiveness of our fine-tuning for multi-view consistent editing, we compare the global consistency of the results and the model's ability to align with the provided reference images before and after fine-tuning.
We use Mip-NeRF 360~\citep{barron2022mip} as the evaluation set, where 10 prompts are applied to each scene for referring editing, producing 20 images per prompt per scene. 
We then compute the DINO similarity 
\cite{oquab2023dinov2} among the generated images for each prompt in each scene and take the average across all scenes, which serves as a measure of cross-view consistency.
As shown in Figure~\ref{fig:flux_finetune} and Table~\ref{tab:mv_finetune}, fine-tuning significantly improves both the global consistency of the edited results and the model’s ability to faithfully follow the editing cues from the reference views.
\input{tinker/Tables/mv_finetune}

\begin{figure*}[tb]
    \centering
    \vspace{-3mm}
    \includegraphics[width=.95\textwidth]{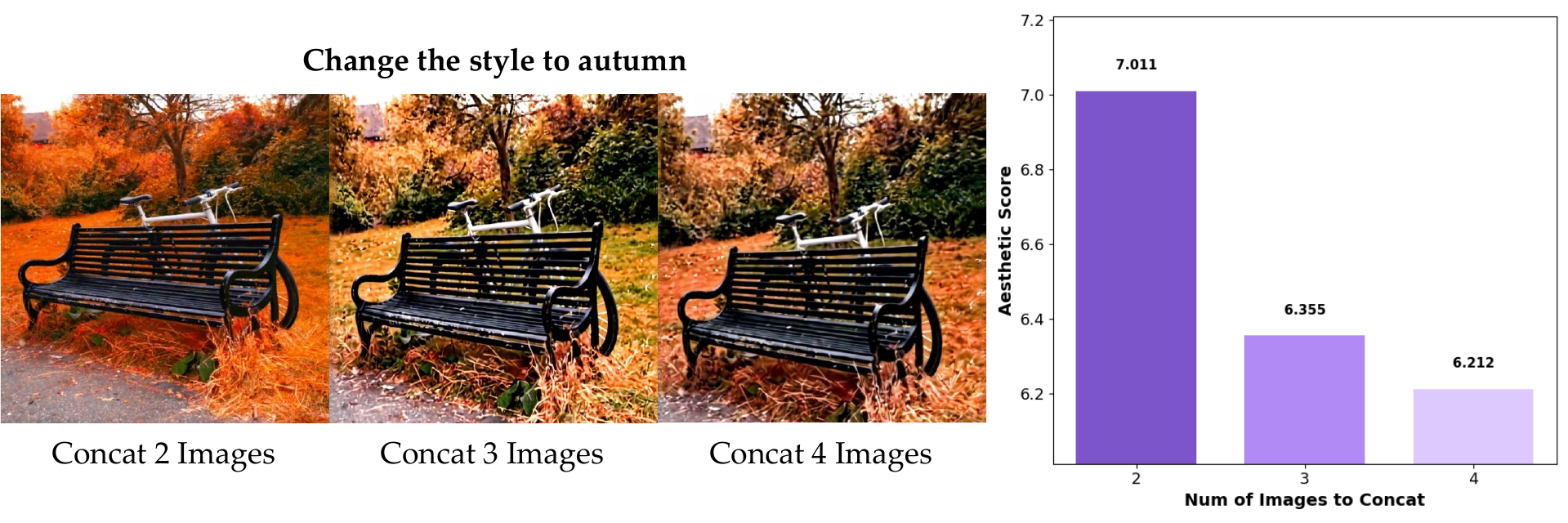}
    \vspace{-5mm}
    \caption{
    Effect of the number of horizontally concatenated images on visual quality. Concatenating too many images leads to a significant degradation in image quality, while concatenating two images yields the best results.
    }
    \label{fig:combine3}
    \vspace{-4mm}
\end{figure*}
\textbf{Concatenating more images for multi-view consistent editing.}
An intuitive approach is to concatenate all multi-view images intended for editing and feed them into the model to achieve consistent editing. However, we observed that this strategy leads to poor results and multi-view inconsistency. Even under a minimal setting, editing with just three concatenated images, the output images exhibit noticeable quality degradation.
This issue arises because the underlying base model imposes a constraint on the image resolution. As a result, input images are automatically resized to fit within the limit. When too many images are concatenated, each individual image is heavily downsampled, leading to substantial loss of detail and visible blurring.
As shown in Figure~\ref{fig:combine3}, we compare the Aesthetic Score across different numbers of concatenated images and find that concatenating two images strikes a reasonable balance between consistency and visual fidelity.

\subsection{Applications}
\label{sec:additional_feature}
\begin{figure*}[htbp]
    \centering
    \vspace{-3mm}
    \includegraphics[width=\textwidth]{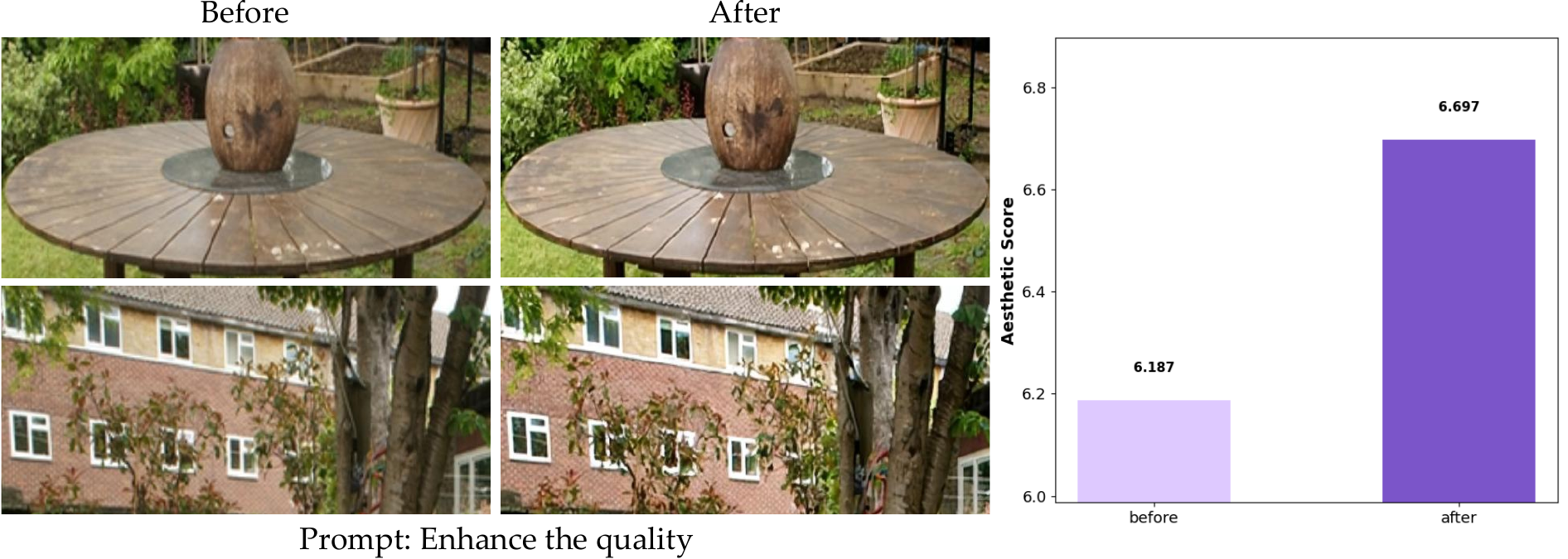}
    \vspace{-5mm}
    \caption{
    \ours\ demonstrates the ability to refine blurry regions, recovering sharper structures and finer details while maintaining overall visual consistency.
    }
    \phantomsection
    \label{fig:enhance_quality}
\end{figure*}

\textbf{Quality refinement.}
We observe that our model can effectively enhance the quality of rendered results, as this refinement can also be regarded as a special variant of editing. As shown in Figure~\ref{fig:enhance_quality}, by employing a prompt such as “enhance the quality,” we can guide the model to refine blurry areas in the rendering, yielding outputs with sharper details. This refinement process allows our method to improve the overall fidelity of the 3DGS reconstruction.

\begin{figure*}[htbp]
    \centering
    \includegraphics[width=\textwidth]{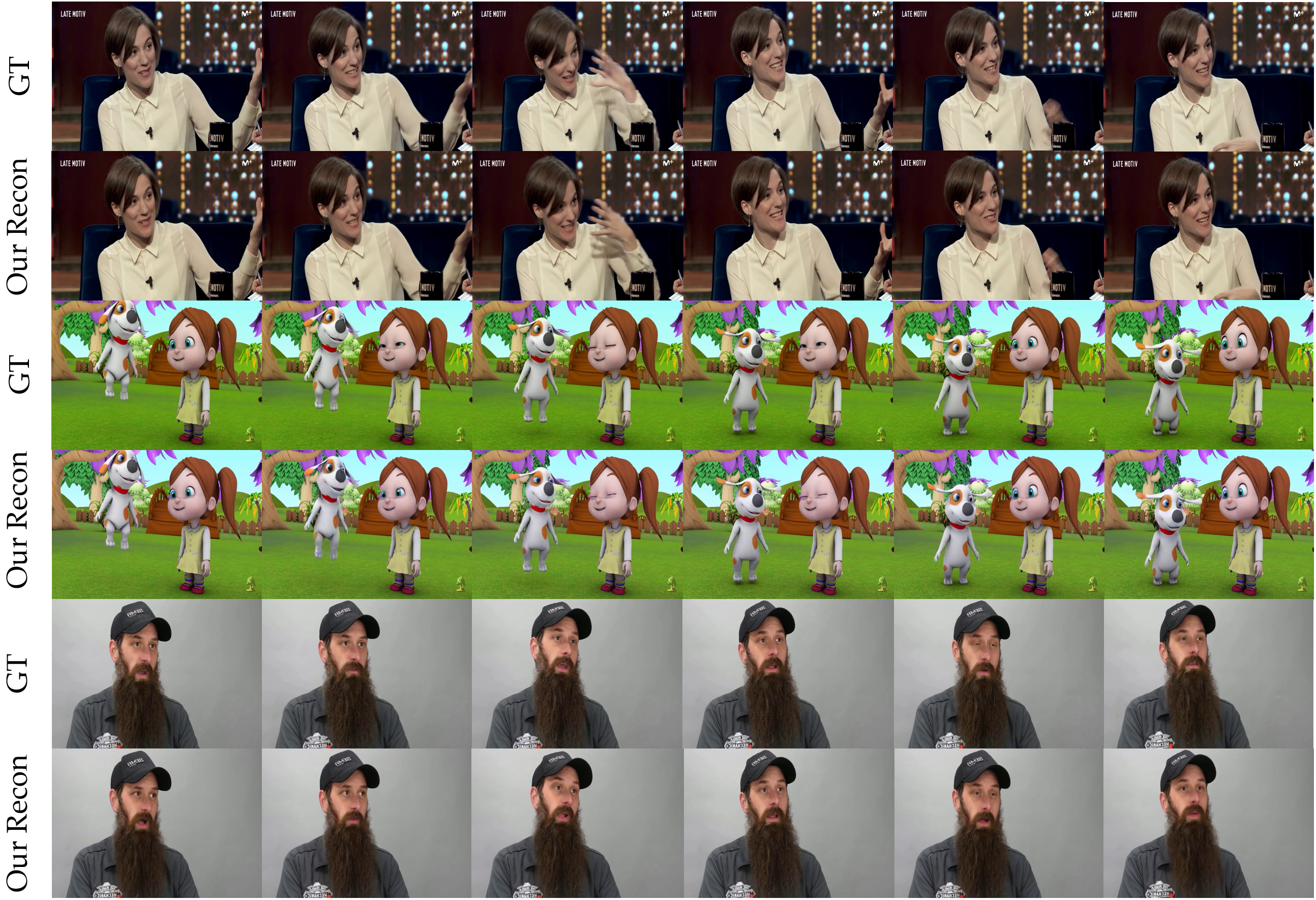}
    \caption{
    \ours\ demonstrates the capability of high-quality video reconstruction with only the first frame and depth maps as input.
    }
    \phantomsection
    \label{fig:video_recon}
\end{figure*}

\textbf{Video reconstruction.}
\ours\ reconstructs high-quality videos from just the first frame and the corresponding depth sequence. As shown in Figure~\ref{fig:video_recon} and Table~\ref{tab:video_recon}, our approach achieves temporally coherent reconstructions with sharp details and faithful geometry, demonstrating its effectiveness across both qualitative and quantitative evaluations. The evaluation dataset consists of 1,000 videos sampled from OpenVid-1M~\citep{nan2024openvid}, and we ensure these videos for evaluation are not included in our training set. We observe that the latest model~\citep{jiang2025vace} taking the first frame and depth as inputs, while capable of producing high-quality video generation, fails to accurately reconstruct the original video content. In contrast, our method significantly improves reconstruction accuracy, achieving faithful recovery of both geometric structure and appearance. Furthermore, because the model operates directly on grayscale depth maps, it naturally supports a compact video representation in which an entire video can be stored as its grayscale depth sequence and a single first frame. This property highlights the potential of our method not only for high-fidelity video reconstruction but also for efficient video compression and storage.
\input{tinker/Tables/video_recon}

\textbf{Test-time optimization.}
Most importantly, since our method does not require per-scene finetuning, it supports test-time optimization: users can iteratively experiment with different reference views, replacing generated views of low quality in the last generation process with newly generated ones using the scene completion model. This iterative process leads to higher-quality 3D editing results.

%% file: tinker/Tables/compare.tex
\begin{table}[bt]
\centering
\caption{
Quantitative comparisons of different methods.
}
\label{tab:compare} 
\begin{tabular}{l
                >{\centering\arraybackslash}p{1.5cm}  
                >{\centering\arraybackslash}p{1.1cm}  
                >{\centering\arraybackslash}p{1.6cm} 
                >{\centering\arraybackslash}p{1.9cm} 
                >{\centering\arraybackslash}p{2.8cm}}
\toprule
 & \multicolumn{3}{c}{\textbf{Quality}} & \multicolumn{2}{c}{\textbf{Computational Cost}} \\ 
\cmidrule(lr){2-4} \cmidrule(lr){5-6}
 & CLIP-dir$\uparrow$ & DINO$\uparrow$ & Aesthetic$\uparrow$ & On 24G GPU & Avg. Editing Time$\downarrow$ \\ 
\midrule
DGE           & 0.102 & 0.948 & 5.747  & Yes & 10min \\
GaussCtrl     & 0.123 & 0.957 & 5.624  & No & 20min \\
TIP-Editor    & 0.084 & 0.875 & 5.397  & No & 35min \\
EditSplat     & 0.102 & 0.956 & 5.661  & Yes & 19min \\ \midrule
Ours-one-shot & 0.143 & 0.958 & 6.214  & Yes & 15min \\
Ours-few-shot & \textbf{0.157} & \textbf{0.959} & \textbf{6.338}  & Yes & 15min \\
\bottomrule
\end{tabular}
\end{table}

%% file: tinker/Tables/mv_finetune.tex
\begin{table}
    \footnotesize
    \setlength\tabcolsep{2.5pt}
    \centering
    \caption{After multi-view consistent image editing fine-tuning, the edited images exhibit substantially improved multi-view consistency, while maintaining comparable text–image alignment and aesthetic quality to the non-finetuned results.}
    \vspace{2mm}
    \label{tab:mv_finetune}
    \resizebox{.486\linewidth}{!}{
    \begin{tabular}{l >{\centering\arraybackslash}p{3cm} >{\centering\arraybackslash}p{2cm}}
        \toprule
        & Before & After \\
        \midrule
        DINO & 0.862 & \textbf{0.943} \\
        CLIP-dir & 0.277 & \textbf{0.281} \\
        Aesthetic & \textbf{7.058} & 6.973 \\
        \bottomrule
    \end{tabular}
    }
    \vspace{-3mm}
\end{table}

%% file: tinker/Tables/video_recon.tex
\begin{table}[h!]
    \footnotesize
    \setlength\tabcolsep{2.5pt}
    \centering
    \caption{Quantitative comparisons of video reconstruction with first frame and depth as input.}
   {
    \begin{tabular}{l >{\centering\arraybackslash}p{3cm} >{\centering\arraybackslash}p{2cm}}
        \toprule
        & PSNR$\uparrow$ & SSIM$\uparrow$ \\
        \midrule
        VACE & 16.635 & 0.331 \\
        \ours & \textbf{31.869} & \textbf{0.941} \\
        \bottomrule
    \end{tabular}
    }
    \phantomsection
    \label{tab:video_recon}
    \vspace{-3mm}
\end{table}

%% file: tinker/Chapters/5_Conclusion.tex
\vspace{-3mm}
\section{Conclusion}
\vspace{-2mm}
We propose \oursbf, to the best of our knowledge, the first general-purpose 3D editing framework that eliminates the need for per-scene optimization. \ours\ bridges a critical gap by extending the breakthroughs of 2D diffusion models into the domain of 3D editing, enabling high-quality results in few-shot even one-shot settings. We also introduce the first dataset and data pipeline specifically designed for multi-view consistent editing to benefit future researches.
Beyond editing, \ours\ also demonstrates additional versatility across tasks such as video compression and video editing, showcasing the potential of a unified 2D and 3D editing framework. We believe that 
\ours\ offers a scalable, flexible, and generalizable solution for future editing research.

%% file: tinker/Appendix/Additional_Implementations.tex
\section{Implementation Details}
\label{append:additional_implementations}

\subsection{Multi-view Consistent Image Editing Dataset and Model}

For the multi-view consistent image editing model, we adopt Flux Kontext~\citep{labs2025flux1kontextflowmatching} as the foundation model and construct a referring multi-view consistent image editing dataset following the procedure detailed in Section~\ref{sec:image_editing}. We use GPT-o3 to generate 400 image editing instructions. The inputs of our data pipeline are sourced from publicly available 3D-aware datasets, including DL3DV~\citep{ling2024dl3dv}, WildRGBD~\citep{xia2024rgbd}, and uCO3D~\citep{liu24uco3d}.
We randomly select two images from each scene, concatenate them, and perform editing. The results are filtered using the procedure described in Section~\ref{sec:image_editing} to determine whether they should be retained.
During dataset construction, two thresholds for data validity, $\tau_{noedit}$ and $\tau_{mv}$, are set quite strictly. Although this may occasionally filter out some good samples, it effectively reduces the number of bad cases in the dataset, thereby benefiting model training. Specifically, we set 
$\tau_{noedit}=0.95$ and $\tau_{mv}=0.9$ to ensure sufficient data quality.
In total, our dataset comprises 250,000 samples, each containing two original images, one edited image, and the corresponding editing instruction. We show the input to Large Vision Language Model to generate the editing prompts in Figure~\ref{fig:prompt} and illustrate some data samples in Figure~\ref{fig:data_sample}. Both the dataset and the data generation pipeline will be released to facilitate further research in this area.

When fine-tuning for referring editing using LoRA~\citep{hu2021lora}, we apply LoRA with rank 128 to all the query, key, value, and output layers of the base model. Training is performed with a dropout rate of 0.05 for 30,000 iterations on four NVIDIA H100 GPUs, using a constant learning rate of 2e-5 and the AdamW optimizer~\citep{loshchilov2017decoupled}.

\subsection{Scene Completion Model}

For scene completion model, we employ Wan2.1 1.3B model~\cite{wan2025wan} as the foundational backbone of our scene completion model. Our model undergoes a two-stage training protocol. Initially, it is pre-trained on the large-scale OpenVid-1M dataset~\citep{nan2024openvid}. Subsequently, to instill robust 3D-aware capabilities, we fine-tuned the model on a curated collection of 3D-centric datasets, including DL3DV~\citep{ling2024dl3dv}, Re10k~\citep{zhou2018stereo}, ArkitScenes~\citep{baruch2021arkitscenes}, WildRGBD~\citep{xia2024rgbd}, and uCO3D~\citep{liu24uco3d}. Depth annotations for our training data are generated using the Video Depth Anything model~\citep{chen2025video}. The training of scene completion model was conducted for 200,000 iterations on a cluster of 16 NVIDIA H100 GPUs using a constant learning rate of 2e-5. The evaluation datasets are from Mip-NeRF-360~\citep{barron2022mip} and IN2N~\citep{haque2023instruct}. Finally, we use NeRFStudio~\citep{tancik2023nerfstudio} for 3DGS optimization and rendering.

\begin{figure*}[htbp]
    \centering
    \vspace{-3mm}
    \includegraphics[width=\textwidth]{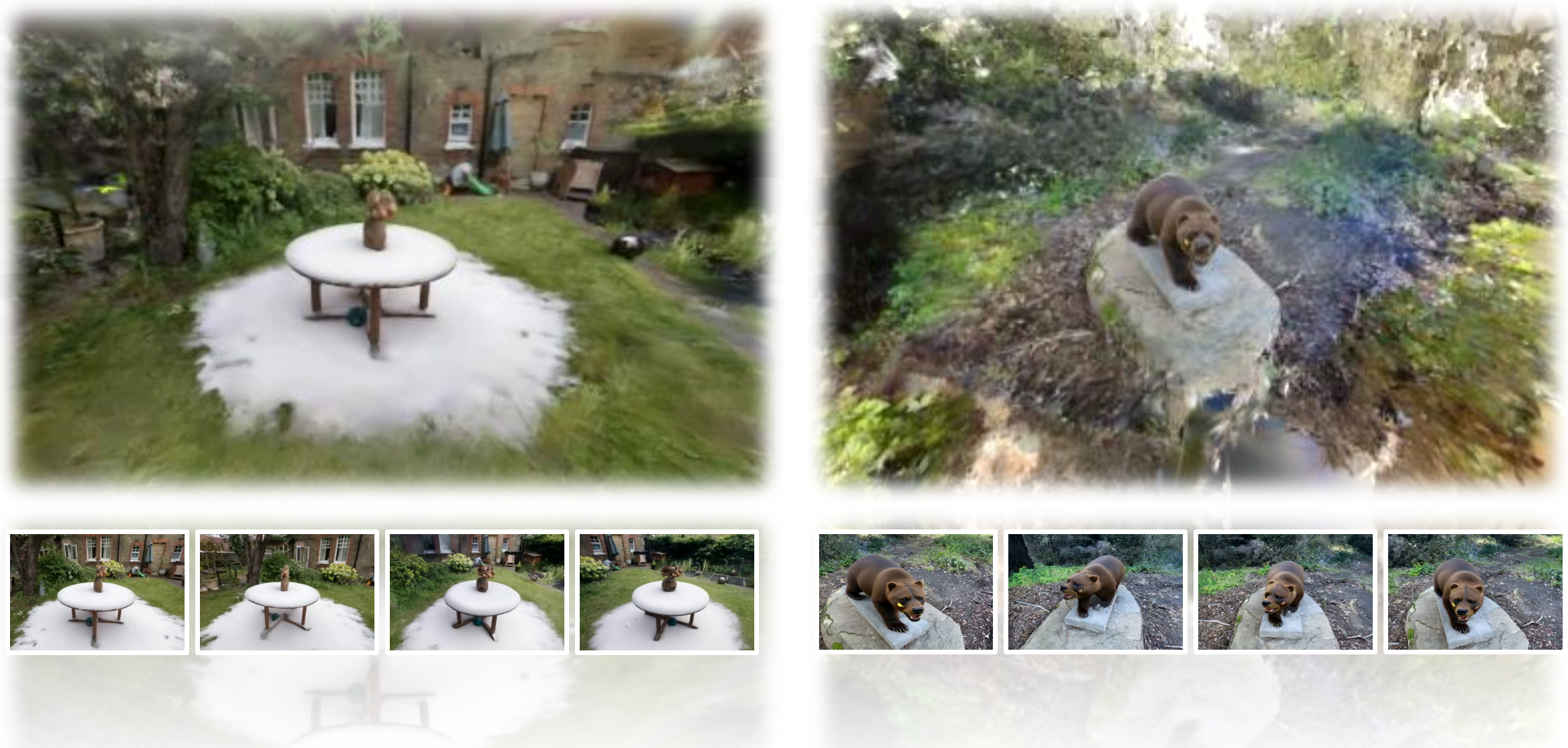}
    \vspace{-5mm}
    \caption{
    Visualizations of edited 3DGS and renderings using NeRFStudio.
    }
    \phantomsection
    \label{fig:3dgs}
    \vspace{-3mm}
\end{figure*}

%% file: tinker/Appendix/Additional_Results.tex
 
\section{Additional Results}

\subsection{Additional Visualizations}
We employ NeRFStudio~\citep{tancik2023nerfstudio} to optimize the 3D Gaussian Splatting using the edited views as input, as illustrated in Figure~\ref{fig:3dgs}. In this section, we further present additional one-shot and few-shot 3D editing results in Figure~\ref{fig:oneshot1}, \ref{fig:oneshot2}, \ref{fig:fewshot1}, \ref{fig:fewshot2}. These comprehensive visualizations demonstrate that our method significantly lowers the usage barrier, enabling high-quality object-level and scene-level  3D editing of various styles without requiring per-scene fine-tuning.

\subsection{Additional Ablations}
\label{append:additional_ablations}

\begin{figure*}[tb]
    \centering
    \includegraphics[width=\textwidth]{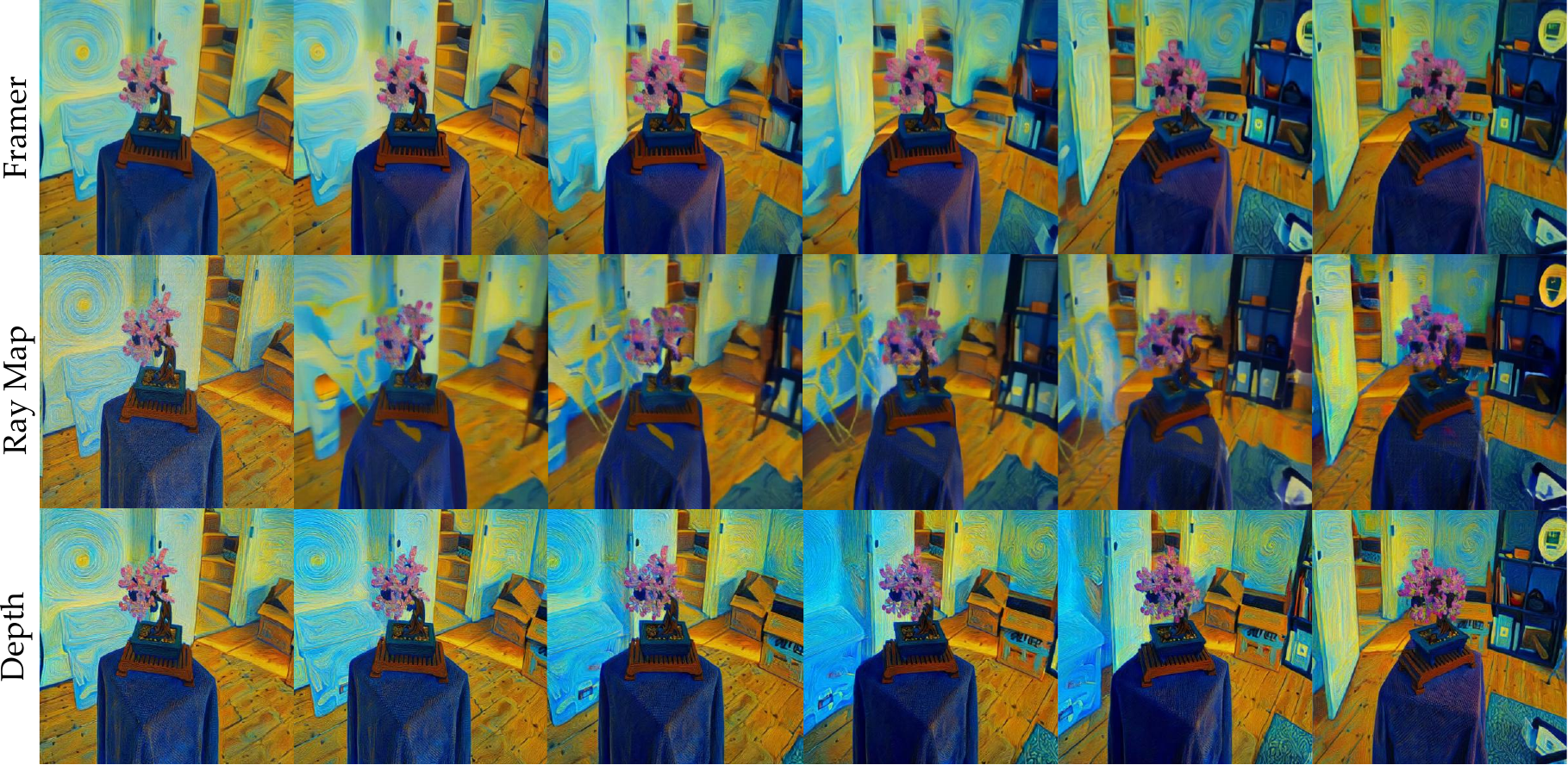}
    \vspace{-4mm}
    \caption{
    Qualitative comparisons of different scene completion methods. Conditioning on depth produces results with one-to-one corresponding camera poses, while achieving superior geometry and detail preservation without being restricted to first and last frame as inputs.
    }
    \label{fig:ray_depth}
    \vspace{-3mm}
\end{figure*}

\textbf{Advantages of depth as condition in scene completion.}
Existing methods predominantly rely on diffusion models to generate new views.
Some approaches~\citep{gao2024cat3d} condition on ray maps to generate missing views, while others~\citep{wang2024framer} directly interpolate between the first and last frames using diffusion models.
We systematically compare our scene completion model with both types of methods to demonstrate its superiority.
First, we train a scene completion model conditioned on ray maps in a manner similar to prior work. We observe that, due to the lack of explicit geometric constraints, this approach often results in noticeable geometric distortions in the generated views, especially in object-level editing setting. Moreover, it often generates outputs that violate the constraints of the camera ray map.
As for the second approach, interpolating between the first and last frames suffers from both visual artifacts and a lack of camera pose information for the generated intermediate views, making them unsuitable for downstream 3DGS optimization. Moreover, this strategy imposes a rigid constraint on input format, limiting the input to the first and last frames and thereby reducing flexibility in editing scenarios.
In contrast, our method leverages depth to provide strong geometric guidance and maintain tight alignment with the corresponding camera poses. It also supports arbitrary reference views, not limited only to the first and last frames. As evidenced by the results in Figure~\ref{fig:ray_depth} and Table~\ref{tab:video_ablation}, our method significantly outperforms the aforementioned baselines in both quality and consistency.

\begin{figure*}[tb]
    \centering
    \vspace{-1mm}
    \includegraphics[width=\textwidth]{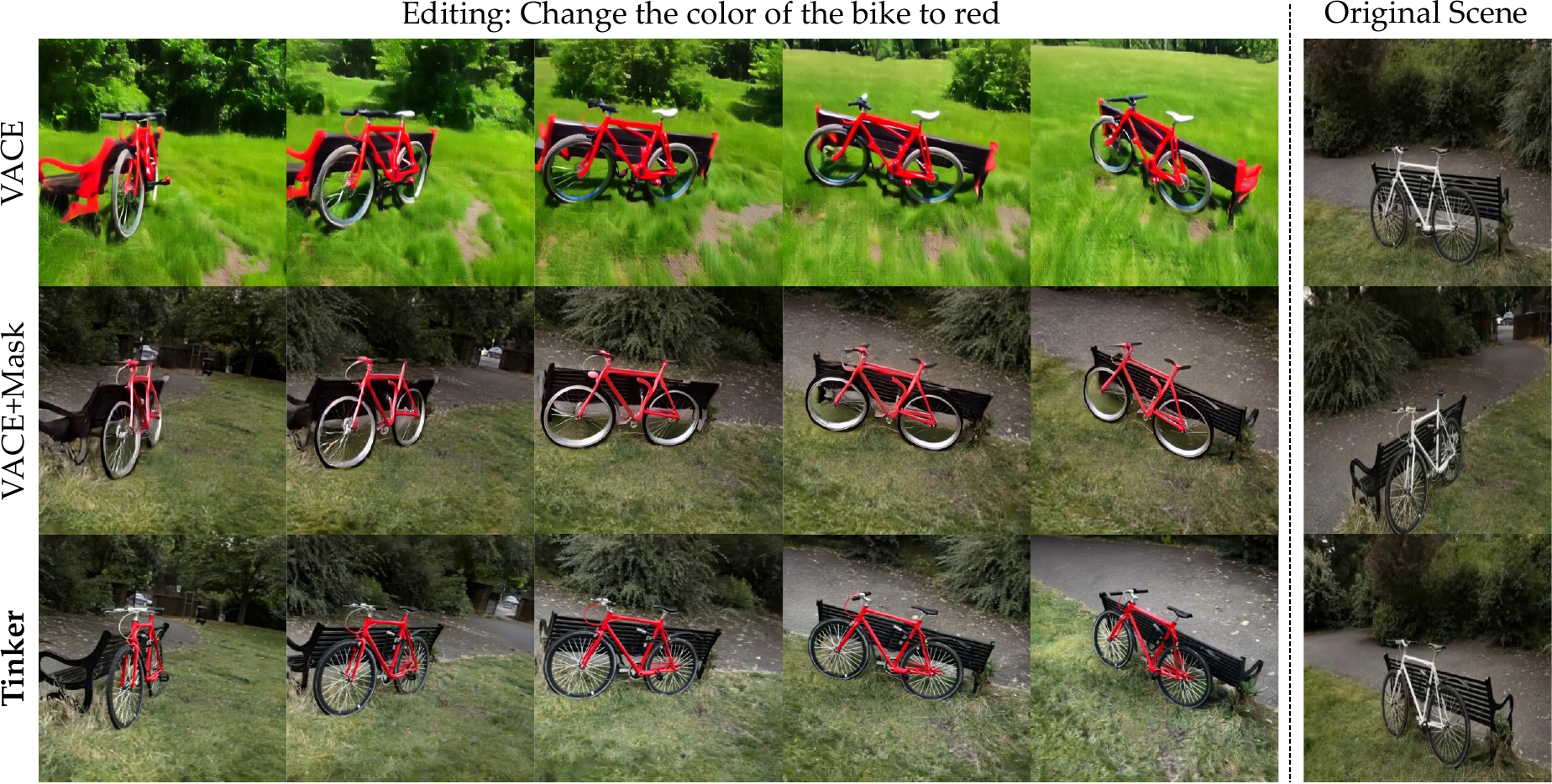}
    \vspace{-5mm}
    \caption{
        Comparison with VACE in both depth-guided video generation and mask-based editing. Our method demonstrates superior multi-view consistency and better preservation of fine details.
    }
    \label{fig:3d_aware}
    \vspace{-3mm}
\end{figure*}

\textbf{Advantages over existing depth-conditioned diffusion models.}
Several recent works, such as VACE~\citep{jiang2025vace}, have explored using depth as condition for video generation. However, these methods are typically trained on natural video datasets and do not pay much attention to 3D-related data. Furthermore, these methods treat the depth condition as a reference rather than a constraint that must be strictly enforced. Consequently, while they are capable of producing high-quality depth-guided video generation, the resulting outputs often fail to strictly adhere to the provided depth constraints, which is not desirable in 3D settings where geometric consistency is essential. 
We compare our approach with the latest existing depth-guided video generation methods VACE~\citep{jiang2025vace} in Figure~\ref{fig:3d_aware} and Table~\ref{tab:video_ablation}, which shows that our model better understands camera motion and more faithfully respects depth constraints. 
In addition to supporting depth-guided video generation, VACE also allows controlling the editing region via masks. We compare our method against both of these capabilities. For depth-guided video generation, the results exhibit clear multi-view inconsistencies. For mask-based editing, the results also show certain fine-grained multi-view inconsistencies, and the quality preservation in detailed regions is significantly lower than that achieved by our method. We attribute this success to removing the text prompt input and training the model on 3D-aware datasets to strictly follow the provided depth.

\input{tinker/Tables/video_ablations}

%% file: tinker/Tables/video_ablations.tex
\begin{table}[htbp]
\centering
\caption{
Quantitative comparisons of different conditions and different depth-guided video generation models. Our approach achieves the best overall performance.
}
\label{tab:video_ablation} 
\begin{tabular}{l
                >{\centering\arraybackslash}p{4cm}  
                >{\centering\arraybackslash}p{1.1cm}  
                >{\centering\arraybackslash}p{1.6cm} 
                }

\toprule

 & Text-Image Similarity$\uparrow$ & DINO$\uparrow$ & Aesthetic$\uparrow$ \\ 
\midrule
VACE         & 0.760 & 0.916 & 5.833  \\
VACE+Mask    & 0.799 & 0.954 & 6.118  \\
Framer       & 0.773 & 0.973 & 6.227  \\
Ours-Ray-Map & 0.783 & 0.931 & 6.214  \\ \midrule
Ours-Depth & \textbf{0.821} & \textbf{0.978} & \textbf{6.586} \\
\bottomrule
\end{tabular}
\end{table}

%% file: tinker/Appendix/Limitations.tex
\section{Limitations and Discussions}
Although our method significantly lowers the barrier to 3D editing, it still has some limitations. First, our dataset is synthesized using the foundation model, which occasionally results in inconsistencies in certain fine details across samples. Second, since our scene completion model operates under depth constraints, it is currently unable to handle edits involving large geometric deformations. We leave these limitations as directions for future work. Nevertheless, despite these issues, we demonstrate strong performance in a wide range of scenarios, offering an effective solution for high-quality, efficient, and user-friendly 3D editing.

\begin{figure*}[htbp]
    \centering
    \vspace{-3mm}
    \includegraphics[width=\textwidth]{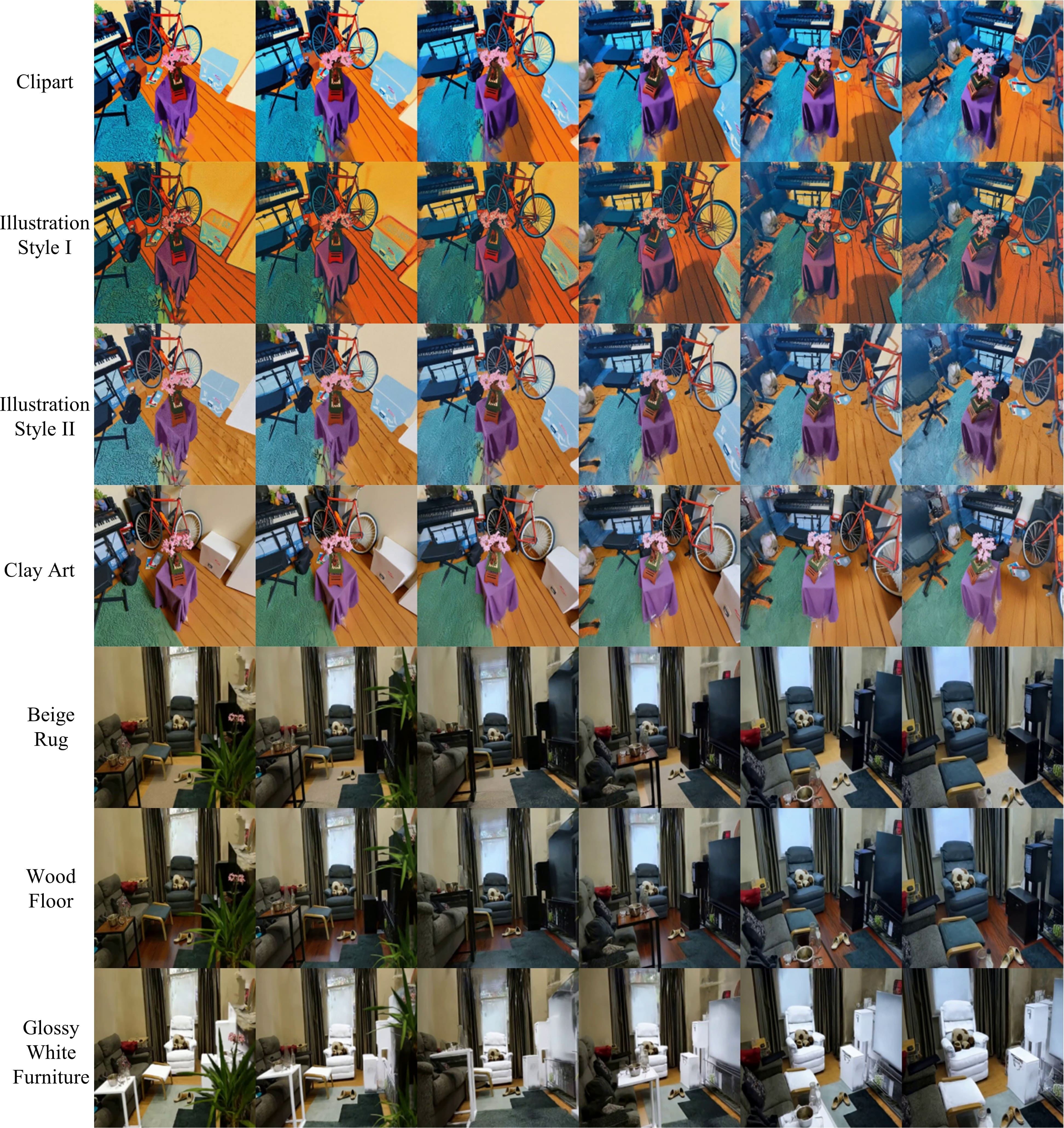}
    \vspace{-5mm}
    \caption{
    \textbf{Additional one-shot editing results without per-scene fine-tuning.}
    }
    \phantomsection
    \label{fig:oneshot1}
    \vspace{-3mm}
\end{figure*}

\begin{figure*}[htbp]
    \centering
    \vspace{-3mm}
    \includegraphics[width=\textwidth]{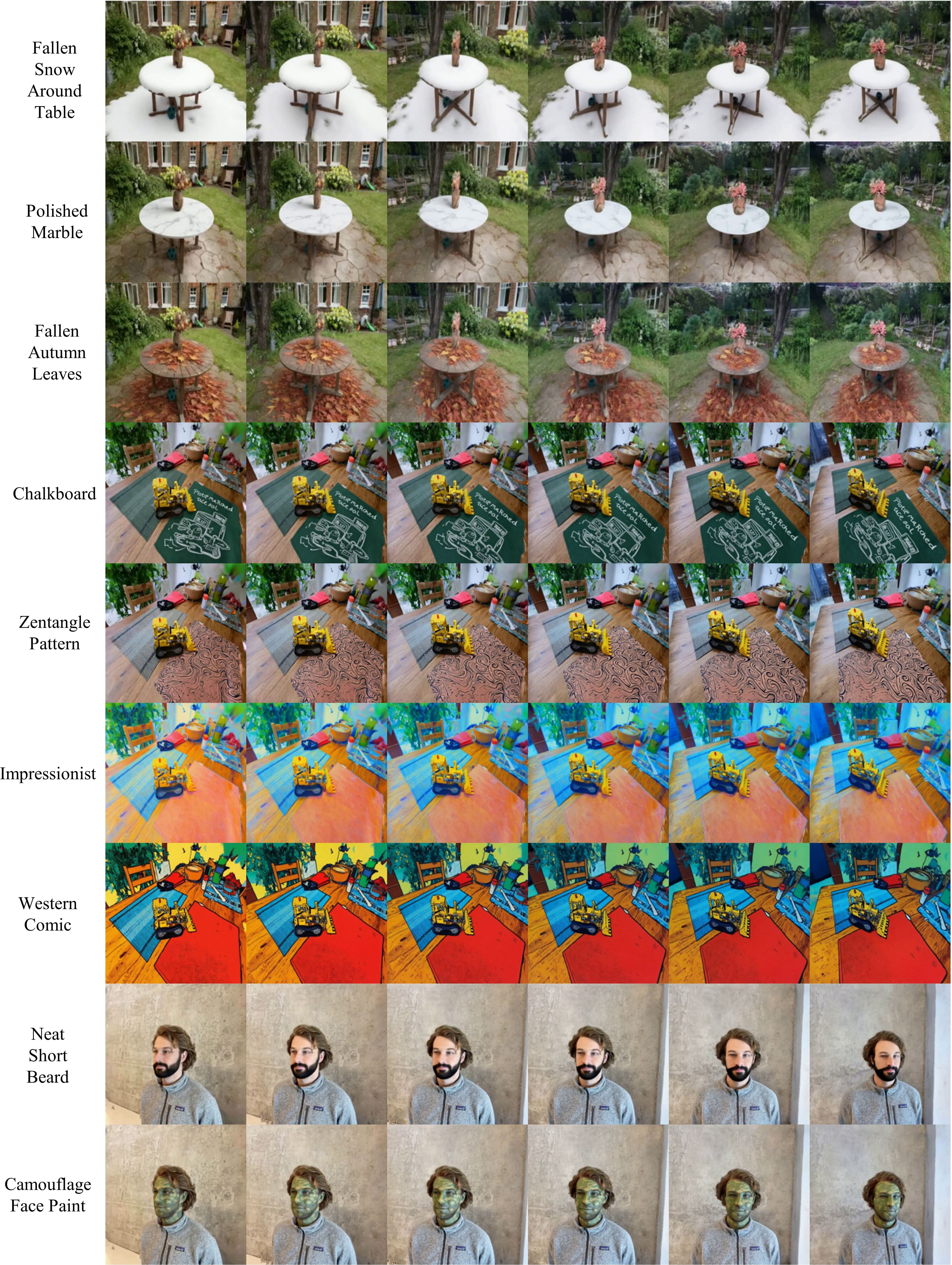}
    \vspace{-5mm}
    \caption{
    \textbf{Additional one-shot editing results without per-scene fine-tuning.}
    }
    \phantomsection
    \label{fig:oneshot2}
    \vspace{-3mm}
\end{figure*}

\begin{figure*}[htbp]
    \centering
    \vspace{-3mm}
    \includegraphics[width=\textwidth]{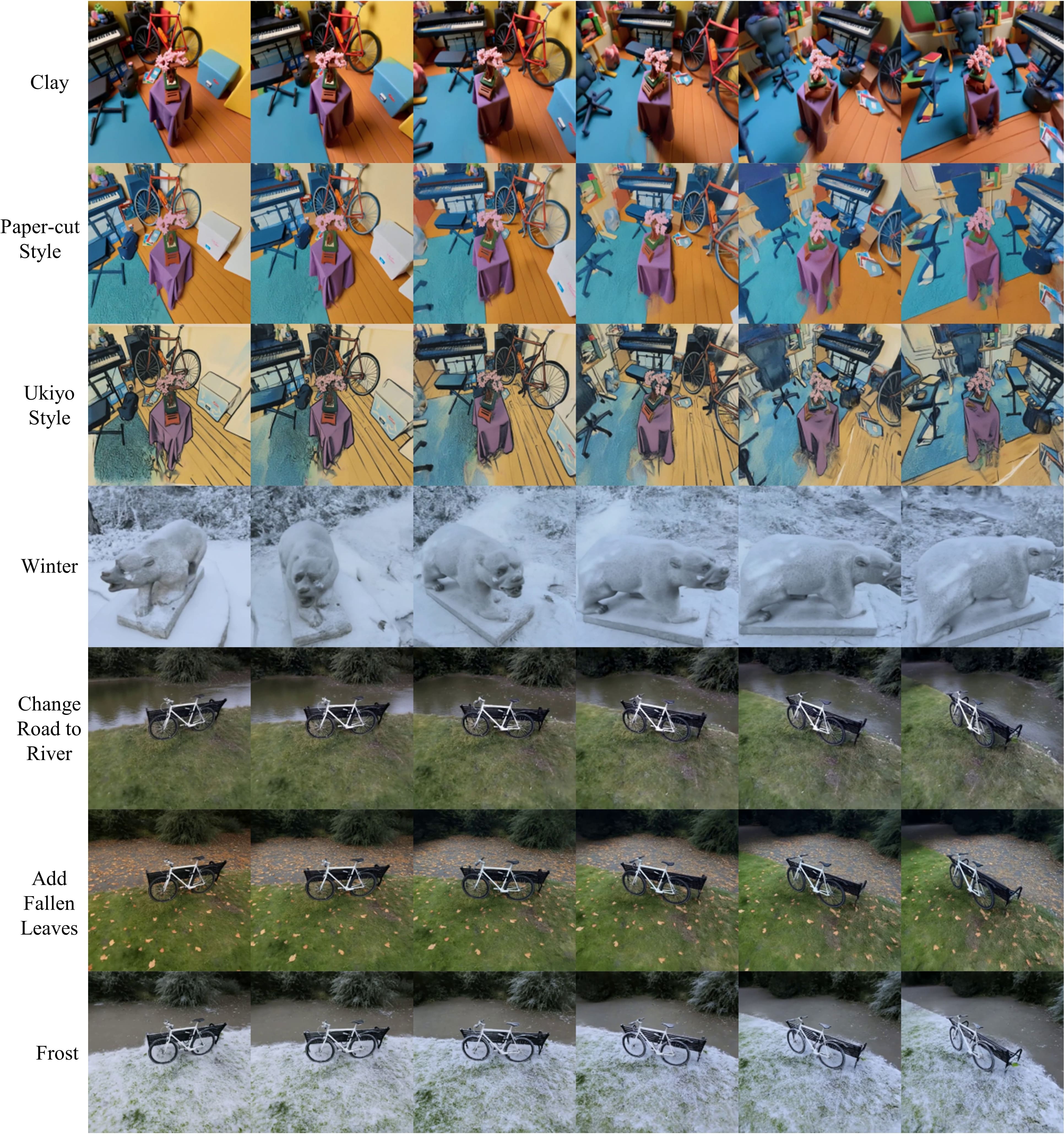}
    \vspace{-5mm}
    \caption{
    \textbf{Additional few-shot editing results without per-scene fine-tuning.}
    }
    \phantomsection
    \label{fig:fewshot1}
    \vspace{-3mm}
\end{figure*}

\begin{figure*}[htbp]
    \centering
    \vspace{-3mm}
    \includegraphics[width=\textwidth]{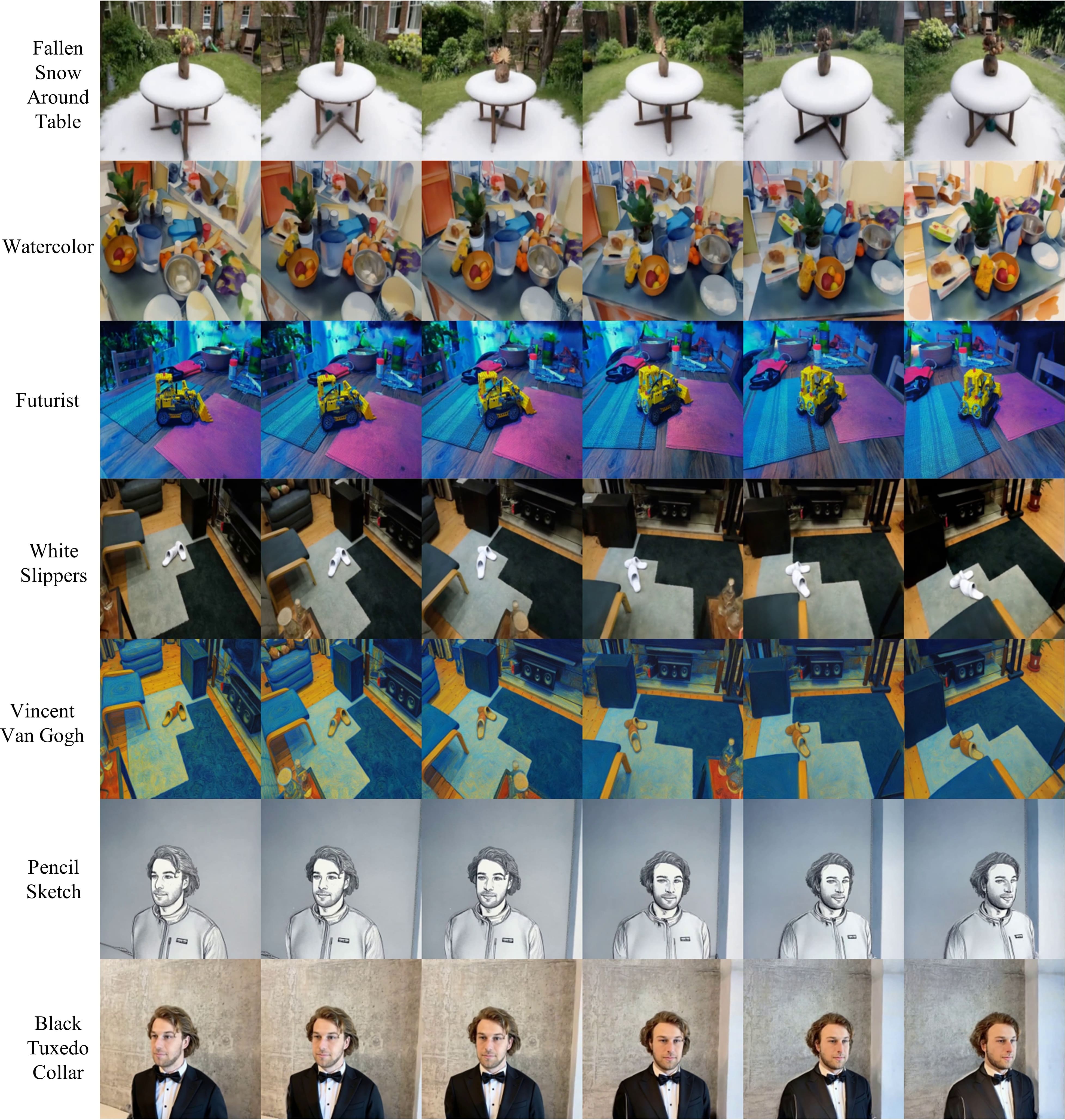}
    \vspace{-5mm}
    \caption{
    \textbf{Additional few-shot editing results without per-scene fine-tuning.}
    }
    \phantomsection
    \label{fig:fewshot2}
    \vspace{-3mm}
\end{figure*}

\begin{figure*}[htbp]
    \centering
    \includegraphics[width=.9\textwidth]{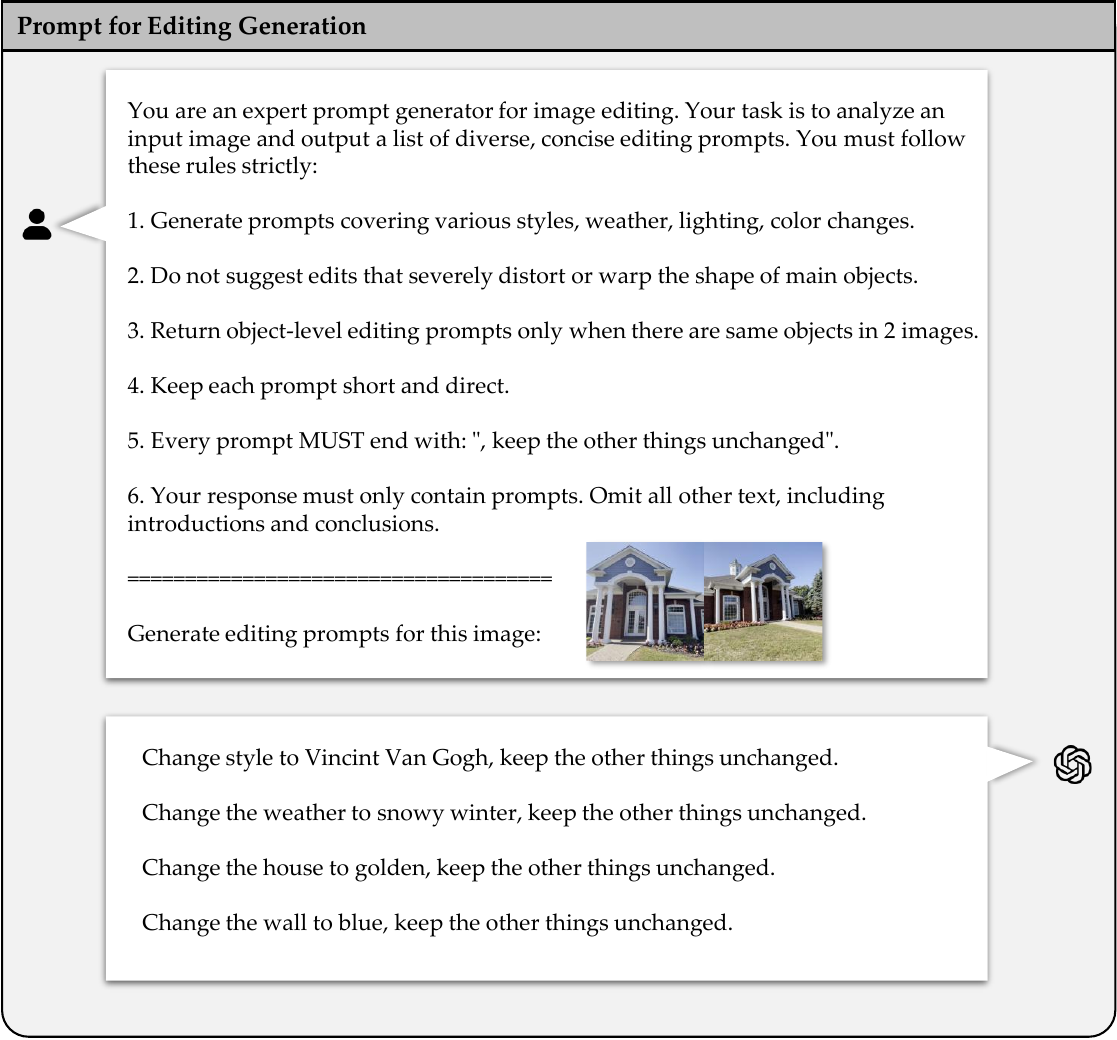}
    \caption{
    Input to a multi-modal large model for the generation of editing prompts.
    }
    \phantomsection
    \label{fig:prompt}
\end{figure*}

\begin{figure*}[htbp]
    \centering
    \includegraphics[width=.9\textwidth]{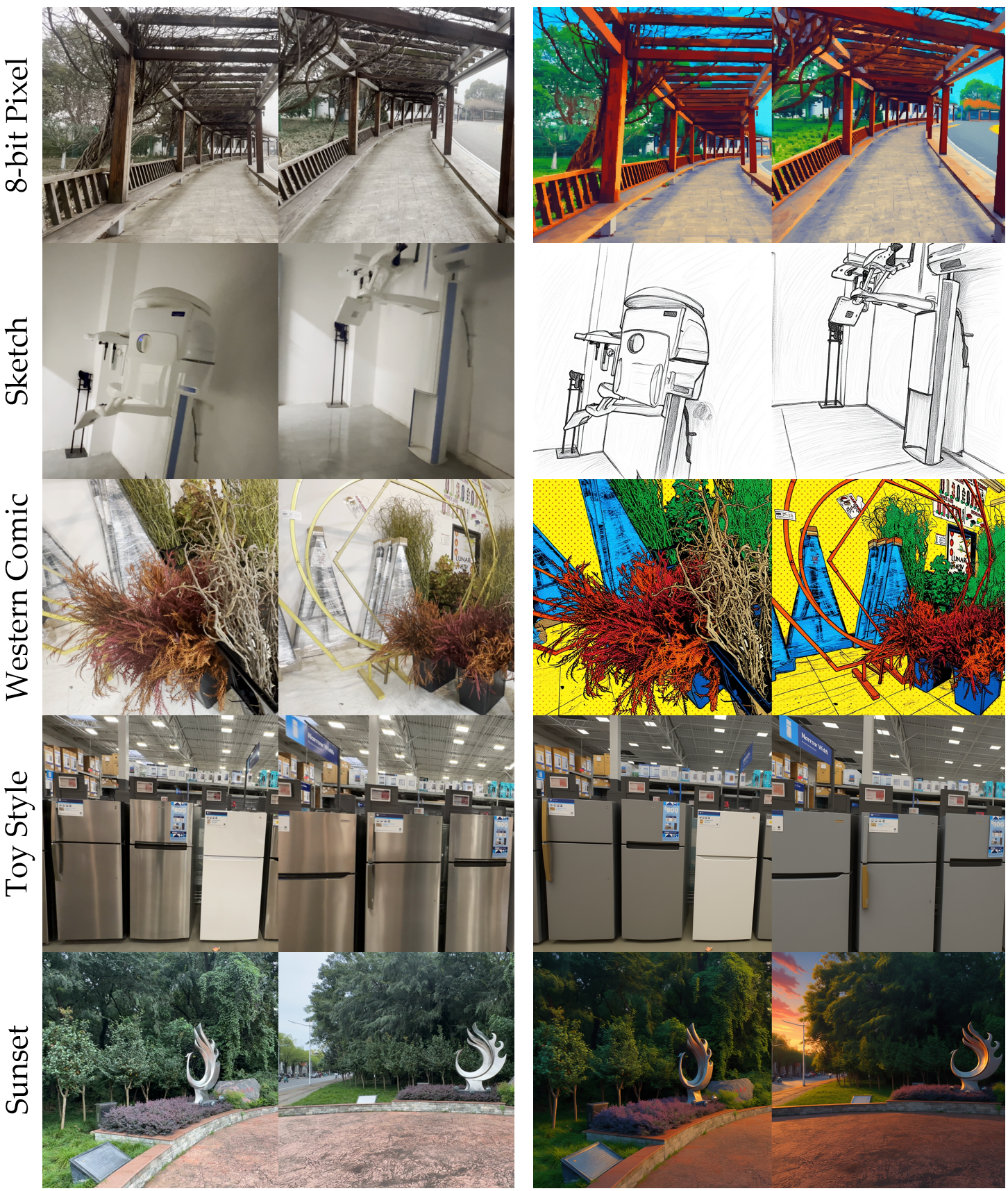}
    \caption{
    Examples from our synthesized multi-view consistent editing dataset. The dataset covers a wide variety of editing, including different weather conditions, lighting setups, and artistic styles.
    }
    \phantomsection
    \label{fig:data_sample}
\end{figure*}